\documentclass[runningheads]{llncs}

\usepackage{graphicx}
\usepackage{mathptmx}
\usepackage{times}
\usepackage{epsfig}
\usepackage{subfigure}
\usepackage{amsmath}
\usepackage{amssymb}
\usepackage{bm}
\usepackage{multirow}
\usepackage[table,xcdraw,dvipsnames]{xcolor}
\usepackage{siunitx}
\usepackage{float}
\usepackage{newfloat}
\usepackage{mathtools}
\usepackage{placeins}
\usepackage[ruled,vlined]{algorithm2e}
\usepackage{booktabs}
\usepackage{appendix}
\usepackage{hyperref}
\usepackage{capt-of}

\DeclareMathAlphabet{\mathcal}{OMS}{cmsy}{m}{n}

\DeclareMathOperator*{\argmax}{argmax}

\DeclarePairedDelimiterX{\infdivx}[2]{(}{)}{%
  #1\;\delimsize\|\;#2%
}
\newcommand{\infdiv}{D\infdivx}
\newcommand{\wsim}{\!\! \sim \!\!}
\newcommand{\mstd}[2]{$#1 {\scriptstyle \pm #2}$}
\newcommand{\bstd}[2]{$\bm{#1 {\scriptstyle \pm #2}}$}
\newcommand{\sm}[1]{{\small #1}}

\begin{document}

\title{An Exploration of Embodied Visual Exploration}

\author{Santhosh K. Ramakrishnan\inst{1,3} \and
Dinesh Jayaraman\inst{2,3} \and
Kristen Grauman\inst{1,3}}

\institute{The University of Texas at Austin \and
University of Pennsylvania \and
Facebook AI Research \\
\email{srama@cs.utexas.edu,~dineshj@seas.upenn.edu,~grauman@cs.utexas.edu}\\
}

\authorrunning{S. Ramakrishnan et al.}

\date{}

\maketitle

\begin{abstract}
Embodied computer vision considers perception for robots in novel, unstructured environments. Of particular importance is the embodied visual exploration problem: how might a robot equipped with a camera scope out a new environment?
Despite the progress thus far, many basic questions pertinent to this problem remain unanswered: 
(i) What does it mean for an agent to explore its environment well?  
(ii) Which methods work well, and under which assumptions and environmental settings? 
(iii) Where do current approaches fall short, and where might future work seek to improve? 
Seeking answers to these questions, we first present a taxonomy for existing visual exploration algorithms and create a standard framework for benchmarking them.
We then perform a thorough empirical study of the four state-of-the-art paradigms using the proposed framework with two photorealistic simulated 3D environments, a state-of-the-art exploration architecture, and diverse evaluation metrics. Our experimental results offer insights and suggest new performance metrics and baselines for future work in visual exploration. Code, models and data are publicly available.~\footnote{Code: \url{https://github.com/facebookresearch/exploring_exploration}}
\end{abstract}

\section{Introduction}

Visual recognition has seen tremendous success in recent years that is driven by large-scale collections of internet data~\cite{ILSVRC15,lin2014microsoft,soomro2012ucf101,kay2017kinetics} and massive parallelization. However, the focus on passively analyzing manually captured photos after learning from curated datasets does not fully address issues faced by real-world robots, who must actively capture their own visual observations. Embodied active perception~\cite{aloimonos1988active,ballard1991animate,bajcsy1988active,wilkes1992active} tackles these problems by learning task-specific motion policies for navigation~\cite{zhu-iccv2017,gupta2017unifying,gupta2017cognitive,savinov2018semi,anderson-eval} and recognition~\cite{malmir2015germs,jayaraman2018pami,yang2019embodied} often via innovations in computer vision and deep reinforcement learning (RL), where the agent is rewarded for reaching a specific target or inferring the right label.

In contrast, in \emph{embodied visual exploration}~\cite{pathak2017curiosity,savinov2018episodic,chen2019learning,ramakrishnan2019emergence}, the goal is inherently more open-ended and task-agnostic: how does an agent learn to move around in an environment to gather information that will be useful for a variety of tasks  that it may have to perform in the future?  

\begin{figure*}
    \centering
    \includegraphics[width=1.0\textwidth]{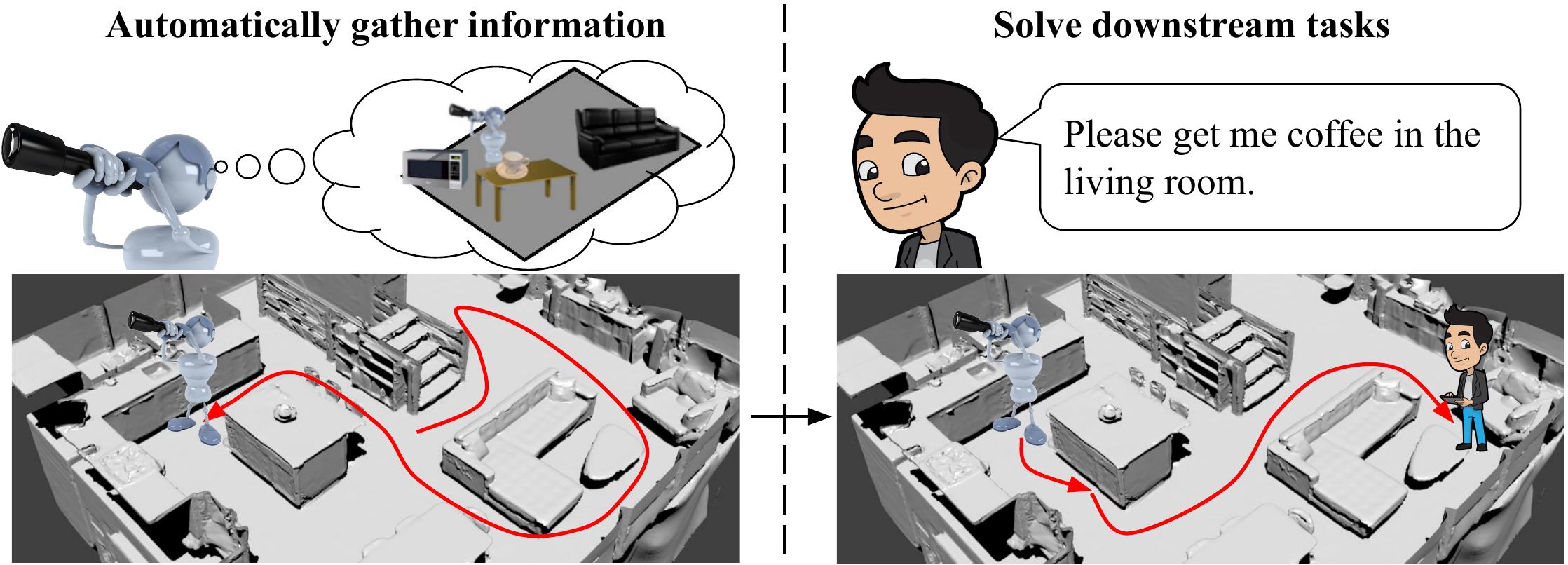}
    \caption{An intelligent agent is dropped into an unknown, unmapped environment. The agent's observations are the egocentric RGB-D views it sees as it moves around.  There is no map provided to the agent; it must look around for obstacles and objects of interest as it moves.  Given a limited time budget, it has to intelligently explore the environment and gather useful information about its geometry and semantics.   Such intelligent exploration prepares an agent to efficiently solve future tasks. For example, if the robot is asked to fetch coffee for a person in the living room, it should use its knowledge of objects, rooms, and the layout of the environment to efficiently perform this task.  This article  organizes and systematically evaluates current paradigms for handling the exploration phase (lefthand panel).}
    \label{fig:intro_figure}
\end{figure*}

Intelligent exploration in 3D environments is important as it allows \emph{unsupervised preparation for future tasks}. Embodied agents that have the ability to explore are flexible to deploy, since they can use knowledge about previously seen environments to quickly gather useful information in the new one, without having to rely on humans. This ability allows them to prepare for unspecified downstream tasks in the new environment. For example, a newly deployed home-robot could prepare itself by automatically discovering rooms, corridors, and objects in the house. After this exploration stage, it could quickly adapt to instructions such as ``Bring coffee from the kitchen to Steve in the living room." See Fig.~\ref{fig:intro_figure}.

A variety of exploration methods have been proposed both in the reinforcement learning and computer vision literature. They employ ideas like curiosity~\cite{schmidhuber1991curious,pathak2017curiosity,burada2018curiosity}, novelty~\cite{strehl2008analysis,bellemare2016unifying,savinov2018episodic}, coverage~\cite{chen2019learning,chaplotmodular}, and reconstruction~\cite{dinesh2018ltla,ramakrishnan2019emergence,seifi2019look} to overcome sparse rewards~\cite{bellemare2016unifying,pathak2017curiosity,burada2018curiosity,savinov2018episodic} or learn task-agnostic policies that generalize to new tasks~\cite{dinesh2018ltla,chen2019learning,ramakrishnan2019emergence}. In this study, we consider algorithms designed to handle complex photorealistic indoor environments where a mobile agent observes the world through its camera's field of view and can exploit common semantic priors from previously seen environments (as opposed to exploration in randomly generated mazes or Atari games). 

How do different exploration algorithms work for different types of environments and downstream tasks?  Despite the growing literature on embodied visual exploration, it has been hard to analyze what works when and why.  This difficulty is due to several reasons. First, prior work evaluates on different simulation environments such as SUN360~\cite{ramakrishnan2019emergence,seifi2019look}, ModelNet~\cite{ramakrishnan2019emergence,seifi2019look}, VizDoom~\cite{pathak2017curiosity,savinov2018episodic}, SUNCG~\cite{chen2019learning}, DeepMindLab~\cite{savinov2018episodic}, and Matterport3D~\cite{chaplotmodular}. Second, prior work uses different baselines with varying implementations, architectures, and reinforcement learning algorithms. Finally, exploration methods have been evaluated from different perspectives such as overcoming sparse rewards~\cite{pathak2017curiosity,burada2018curiosity}, pixelwise reconstruction of environments~\cite{dinesh2018ltla,ramakrishnan2018sidekick,ramakrishnan2019emergence,seifi2019look}, area covered in the environment~\cite{chen2019learning,chaplotmodular}, object interactions~\cite{pathak2018beyond,haber2018selfaware}, or as an information gathering phase to solve downstream tasks such as navigation~\cite{chen2019learning}, recognition~\cite{dinesh2018ltla,ramakrishnan2019emergence,seifi2019look}, or pose estimation~\cite{ramakrishnan2019emergence}. Due to this lack of standardization, it is hard to compare any two methods in the literature.

This paper presents a unified view of exploration algorithms for visually rich 3D environments, and a common evaluation framework to understand their strengths and weaknesses. First, we formally define the exploration task. Next, we provide a taxonomy of exploration approaches and sample representative methods for evaluation. We evaluate these methods and several baselines on common ground: two well-established 3D  datasets~\cite{ammirato2016avd,chang2017matterport} and a state-of-the-art neural architecture~\cite{chen2019learning}. Unlike navigation and recognition, which have clear-cut success measures, exploration is more open-ended. Hence, we quantify exploration quality along multiple meaningful dimensions such as mapping, robustness to sensor noise, and relevance for different downstream tasks. Finally, by systematically evaluating different exploration methods using this benchmark, we highlight their strengths and weaknesses, and we identify key factors for learning good exploration policies.

Our main contribution is the first unified and systematic empirical study of exploration paradigms on 3D environments. In the spirit of recent influential studies in other domains~\cite{duan2016benchmarking,mahmood2018benchmarking,goyal2019scaling,mishkin2019benchmarking,habitat19iccv}, we aim first and foremost to provide a reliable, unbiased, and thorough view of the state of the art that can be a reference to the field moving forward. To facilitate this, we introduce technical improvements to shore up existing approaches, and we propose new baselines and performance metrics. We will publicly release all code to standardize the development and evaluation of exploration algorithms.

Next, we detail the empirical study framework in Sec.~\ref{sec:study_framework}, the taxonomy of existing exploration paradigms in Sec.~\ref{sec:taxonomy}, and the exploration evaluation framework in Sec.~\ref{sec:evaluation_framework}. Finally, we present the experiments and ablation studies in Sec.~\ref{sec:experiments} and conclude in Sec.~\ref{sec:conclusions}.
\section{Empirical study framework}
\label{sec:study_framework}
First, we define a framework for systematically studying different exploration algorithms. We formulate exploration as a sequential information-gathering task: at each time step, the agent takes as input the current view and history of accumulated views, updates its internal model of the environment (e.g., a spatial memory or map), and selects the next action (i.e., camera motion) to maximize the information it gathers about the environment. How the latter is defined is specific to the exploration algorithm, as we will detail in Sec.~\ref{sec:taxonomy}. This process may be formalized as a finite-horizon partially observed Markov decision process (POMDP). In the next few sections, we describe the POMDP formulation, the 3D simulators used to realize it, the policy architecture that accounts for partial observability while taking actions, and the policy learning algorithm.

\subsection{The exploration POMDP}
We model exploration as a partially observed Markov decision process (POMDP)~\cite{lovejoy1991survey}. A POMDP can be represented as a tuple $(\mathcal{S}, \mathcal{O}, \mathcal{A}, \Lambda, \mathcal{T}, \mathcal{R}, \rho_{0}, \mathcal{\gamma}, T_{exp})$ with state space $\mathcal{S}$, observation space $\mathcal{O}$, action space $\mathcal{A}$, state-conditioned observation distribution $\Lambda$, transition distribution $\mathcal{T}$, reward function $\mathcal{R}$, initial state distribution $\rho_{0}$, and  discount factor $\gamma$. The agent is spawned at an initial state $s_{0}\wsim\rho_{0}$ in an unknown environment. At time $t$, the agent at state $s_{t}$ receives an observation $o_{t}\wsim\Lambda(. | s_{t})$, executes a camera motion action $a_{t}\wsim\pi(\hat{s}_{t})$, receives a reward $r_{t}\wsim \mathcal{R}(.|s_{t}, a_{t}, s_{t+1})$, and reaches state $s_{t+1}\wsim \mathcal{T}(. | s_{t}, a_{t})$. The state representation $\hat{s}_{t}$ is obtained using the history of observations $\{o_{k}\}_{k=1}^{t}$, and $\pi$ is the agent's policy. We assume a finite episode length $T_{exp}$. 

The goal is to learn an optimal exploration policy $\pi^{*}$ that maximizes the expected cumulative sum of the discounted rewards over an episode:
\vspace*{-0.05in}
\begin{equation}
    \pi^{*} = \argmax_{\pi} E_{\tau\sim\pi} \bigg[\sum_{t=0}^{T_{exp}} \gamma^{t} r_{t}\bigg],
    \vspace*{-0.025in}
\end{equation}
where $\tau$ is a sequence of $\{(s_{t}, a_{t}, r_{t})\}_{t=1}^{T}$ tuples generated by starting at $s_{0}\sim \rho_{0}$ and behaving according to policy $\pi$ at each time step. The reward function $\mathcal{R}$ captures the method-specific incentive for exploration (see Sec.~\ref{sec:taxonomy}). Next, we concretely define the instantiation of the POMDP in terms of photorealistic 3D simulators.

\begin{figure}[t]
    \centering
    \includegraphics[width=1.00\textwidth,trim={0 0 0 0},clip]{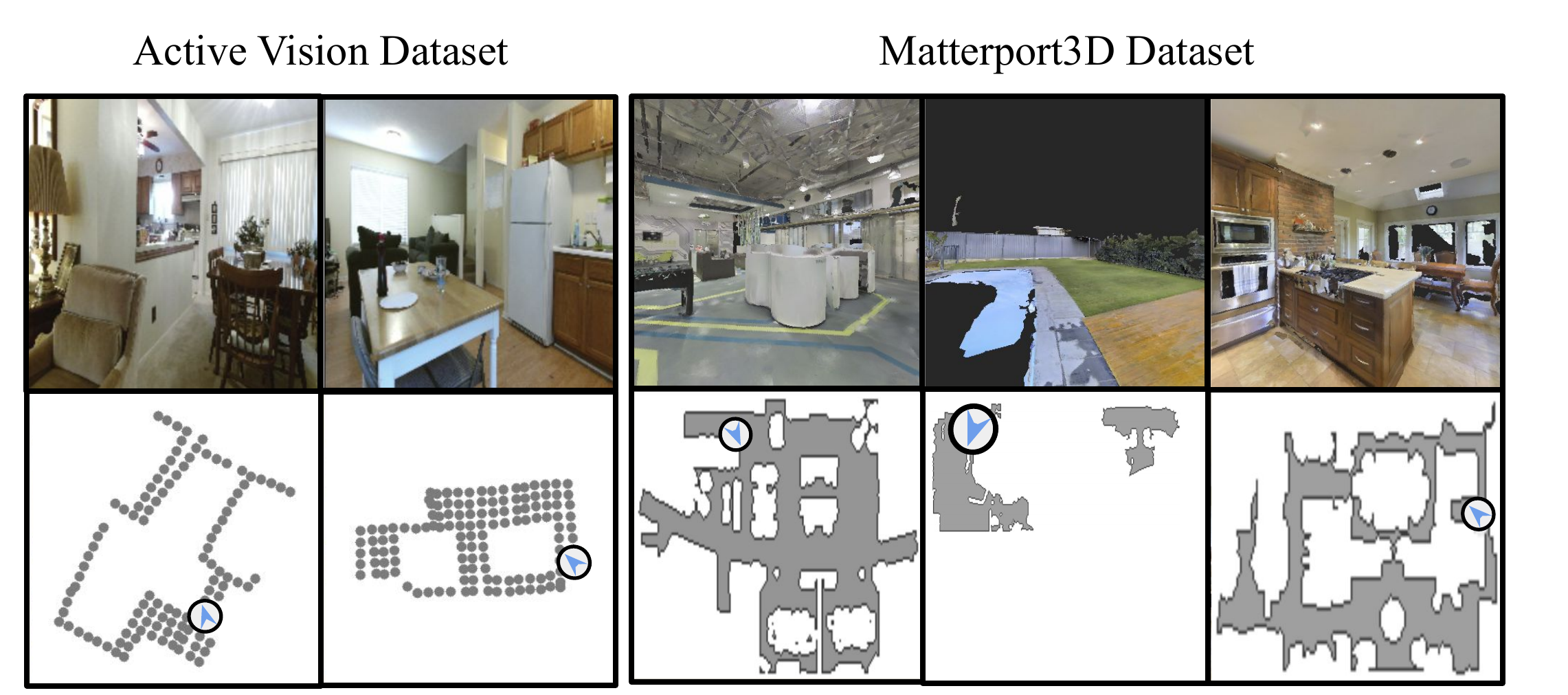}
    \caption{Examples of 3D environment layouts in the Active Vision Dataset~\cite{ammirato2016avd} (first two columns) and Matterport3D~\cite{chang2017matterport} (last three columns). The top row shows a first-person view and the bottom row shows the 3D layout of the environment with free space in gray, occupied space in white, and the viewpoint as the blue arrow.}  
    \label{fig:3d_environments}
\end{figure}

\subsection{Simulators for embodied perception}
\label{sec:datasets}
In order to standardize the experimentation pipeline, we use simulators built on top of two realistic 3D datasets: Active Vision Dataset~\cite{ammirato2016avd} (AVD) and Matterport3D~\cite{chang2017matterport} (MP3D). See Fig.~\ref{fig:3d_environments}. There are several other valuable 3D assets in the community~\cite{xia2018gibson,ai2thor,stanford2d3d,replica19arxiv}; we choose MP3D and AVD for this study due to their complementary attributes (see Tab.~\ref{tab:datasets}).  We anticipate that the different facets of the two datasets will allow us to better elicit any differences in behavior between the paradigms we test.

\begin{table}[t]
\centering
\resizebox{0.8\textwidth}{!}{
\begin{tabular}{@{}lcc@{}}
\toprule
Properties                      & Active Vision Dataset~\cite{ammirato2016avd}    & Matterport3D~\cite{chang2017matterport} \\
\midrule      
View sampling                   & Discrete                                        & Continuous                              \\
Size (average navigable space)  & Small ($9.7\si{m^2}$)                           & Large ($516.34\si{m^2}$)                \\
Train / val / test splits       & 9 / 2 / 4                                       & 61 / 11 / 18                            \\
Large scale data                & No                                              & Yes                                     \\
Outdoor components              & No                                              & Yes                                     \\
Clutter                         & Significant                                     & Mild                                    \\
\midrule                                                    
Forward motion                  & 0.3 $\si{m}$                                    & 0.25 $\si{m}$                           \\
Rotation angle                  & $30^{\circ}$                                    & $10^{\circ}$                            \\
\bottomrule
\vspace{0.05cm} 
\end{tabular}
}
\caption{The contrasting properties of AVD and MP3D provide diverse testing conditions. Last two rows show the action magnitudes.}
\label{tab:datasets}
\end{table}

The \textbf{Active Vision Dataset} (AVD) is a dataset of dense RGB-D scans from 15 unique indoor houses and office buildings~\cite{ammirato2016avd}. We simulate embodied motion by converting the dataset into a connectivity graph with discrete points and moving along the edges of the graph (similar to~\cite{mattersim}). Compared to Matterport3D, AVD environments are relatively smaller in size, fewer in number, and have noisy depth inputs and coarsely discretized movements. The coarse sampling of views in AVD prevents continuous simulation of camera motion, as the reconstructed meshes tend to have significant mesh defects. However, movement along discrete nodes placed approximately $0.3\si{m}$ apart allows a realistic simulation for an embodied agent in cluttered home interiors, which are lacking in real estate photos or computer graphics datasets. 

\textbf{Matterport3D} (MP3D) is a dataset of photorealistic 3D meshes from 90 indoor buildings~\cite{chang2017matterport}. The MP3D environments are generally larger in size, greater in quantity, and offer more finely discretized movements than AVD. Some MP3D environments contain outdoor components such as swimming pools, porches, and gardens which present unique challenges to exploration, as we will see in Sec.~\ref{sec:analysis}. We use the publicly available Habitat simulator~\cite{habitat19iccv}, which provides fast simulation.

In these simulation environments, the state space $\mathcal{S}$ consists of the agent's position and orientation within the environment. The observation space $\mathcal{O}$ consists of an RGB-D sensor, an odometer to track the camera position, and a bump sensor to detect collisions. We consider exploration under both idealized noise-free sensing and realistic noisy conditions. The action space $\mathcal{A}$ is discrete and has three actions: move forward, turn left, and turn right. The motion values are given in Tab.~\ref{tab:datasets}.

We stress that though the agents we test are simulated robots, the 3D visual spaces in which we test them are real---they are comprised of real meshes captured from real images, not generated from models with computer graphics.  While real robots do introduce challenges beyond perception, realistic simulations have the advantage of reproducibility and scaling agent experience, and hence are widely used today~\cite{anderson2018evaluation,chang2017matterport,embodiedqa,habitat19iccv,savva2017minos,replica19arxiv,xia2018gibson}. Furthermore, our experiments examine the effects of noisy odometry and noisy occupancy in these environments.  Finally, recent work shows promising results transferring from simulated environments (including Matterport3D) to real-world exploration and navigation, supporting simulators as a meaningful testbed for our benchmark~\cite{chaplotmodular,kadian2019we}.

\subsection{Policy architecture}
\label{sec:policy}

In order to successfully explore an unknown environment, the agent needs to keep track of where it is currently located, how much of the environment has been explored, what information was contained in the explored regions, and what are the potentially unexplored parts of the environment.  We select a state-of-the-art policy architecture that is well-suited to these objectives and has shown successful exploration in partially-observed and visually rich 3D environments~\cite{chen2019learning}. 

\begin{figure}[t]
    \centering
    \includegraphics[width=1.0\textwidth]{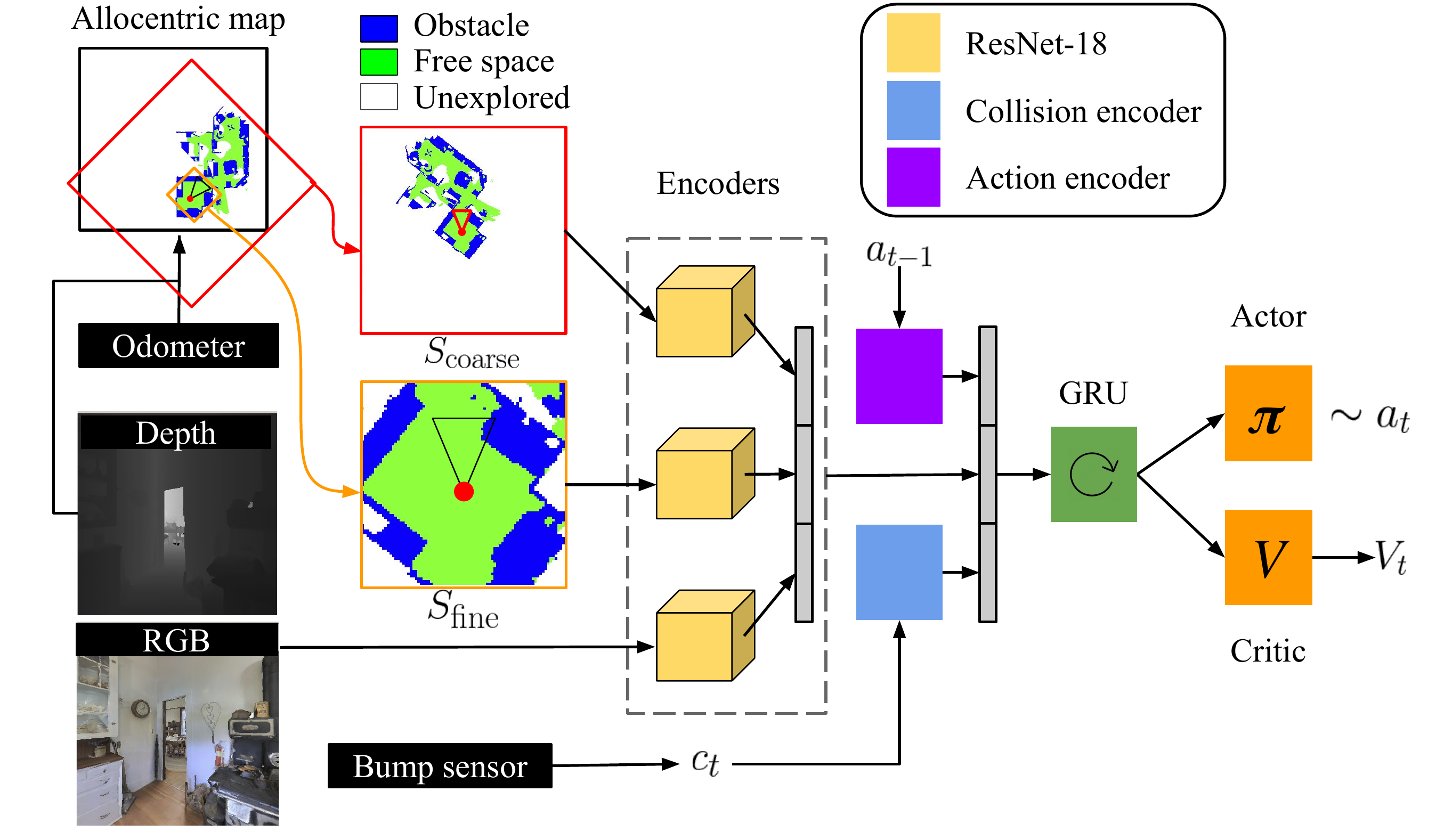}
    \caption{\textbf{Policy architecture:} The agent receives RGB-D, collision, and odometer inputs from the environment at each time-step. The agent maintains a spatial occupancy memory (allocentric map) that is constructed from the depth and pose inputs over time. Local egocentric crops centered around the agent are extracted (orange and red boxes, top left) and encoded along with the RGB input using ResNet encoders. The encoded information is aggregated over time by a temporal recurrent memory (GRU) that additionally encodes the semantics of the environment. The aggregated state of the temporal memory is used by the Actor module to select actions that move the agent around the environment.}
    \label{fig:policy_architecture}
\end{figure}

The agent receives sensory readings from an RGB-D sensor, an odometer that measures the motion resulting from the past action, and a bump sensor that detects collisions resulting from past actions. The policy architecture consists of three parts: (1) spatial memory, (2) temporal memory, and (3) an actor-critic model (see Fig.~\ref{fig:policy_architecture}). We elaborate on these parts in the following.

\vspace{-0.1in}
\paragraph{Spatial memory}
As the agent explores an environment, it needs to keep track of what areas are explored and where the obstacles are present, and then decide where to navigate next in order to explore further. This is done by maintaining a top-down occupancy map that represents free, occupied, and unexplored regions in the environment (see Fig.~\ref{fig:policy_architecture}). The map is built geometrically by using depth inputs to estimate the occupied and free regions based on pixel heights at a given location, and then registering them to the allocentric map based on the agent's pose.

Let $\{(D_{i}, p_{i})\}_{i=1}^{N}$ be the sequence of depth maps $D_{i}$ and pose estimates $p_{i}$ of the agent up to time $N$. The agent's pose at time $i$ is expressed as $p_{i} = (\bm{R}_{i}, \bm{t}_{i})$, with $\bm{R}_{i}, \bm{t}_{i}$ representing the \textit{agent's} camera rotation and translation in the world coordinates. Let the agent initially start at the origin of the map, i.e., $p_{0} = \bm{0}$, and let $\bm{K} \in \mathbb{R}^{3\times 3}$ be the intrinsic camera matrix, which is assumed to be provided. Each pixel $x_{ij}$ in the depth map $D_{i}$ can then be projected to the 3D point cloud as follows:

\begin{equation}
    w_{ij} = \begin{bmatrix}
                \bm{R}_{i} & \bm{t}_{i}\\
                \bm{0} & 1
             \end{bmatrix} \bm{K}^{-1} x_{ij},~\forall j \in \{1, ..., S_i\},~\forall i \in \{1, ..., N\},
\end{equation}

\noindent where $S_i$ is the total number of pixels in $D_{i}$. This gives us the set of points $\bm{W} = \bigcup_{i=1}^{N} \bigcup_{j=1}^{S_i} w_{ij}$ representing all observed points during exploration. We discretize the top-down view of the environment into discrete 2D cells of size $s$ and classify each cell $b$ into three categories---free space, obstacle, and unexplored space---as follows:

\begin{equation}
\textrm{class}(b) = \begin{cases} 
                     \textrm{obstacle}, & \textrm{if}~\exists w \in \bm{W}_{b} \mid \textrm{height}(w) \in [\eta_{l}, \eta_{h}] \\
                     \textrm{free space}, & \textrm{if}~\textrm{height}(w) < \eta_{l}~~\forall~~w \in \bm{W}_{b} \\
                     \textrm{unexplored}, & \textrm{if}~\bm{W}_{b} = \varnothing \\
                     \end{cases}
\end{equation}
where $\textrm{height}(w)$ is the height of point $w$, $\eta_{l}$ and $\eta_{h}$ are height thresholds in meters used to determine occupancy, and $\bm{W}_{b} \subset \bm{W}$ is the set of points present within cell $b$ with heights less than $\eta_{h}$. This gives us the agent's allocentric spatial memory specifying which regions are free space, obstacle, and unexplored space. Based on the allocentric memory and the agent's current pose, egocentric crops at two resolutions $S_{\textrm{coarse}} \times S_{\textrm{coarse}}$ and $S_{\textrm{fine}} \times S_{\textrm{fine}}$ are provided to the next stages, where $S_{\textrm{coarse}}$ and $S_{\textrm{fine}}$ are measured in meters.

\vspace{-0.1in}
\paragraph{Temporal memory}
While the spatial memory purely encodes geometric information, the agent needs to additionally store other textural and semantic cues in the environment over time. This is done via a temporal memory that jointly encodes all sensor inputs and aggregates them over time. 

The RGB and the coarse and fine occupancy maps are encoded as 128-D feature vectors via independent pre-trained ResNet-18~\cite{he2016deep} models, then concatenated and fused into a 512-D feature vector using a fully connected (FC) layer. The previous action and binary collision signal are encoded as learned 32-D embeddings. The encoded image, action, and collision features are concatenated, and then fused using an FC layer to obtain a 512-D feature vector at each time-step. This is temporally aggregated using a Gated Recurrent Unit (GRU)~\cite{chung2014empirical} with one hidden-layer and 512-D hidden-state.

\vspace{-0.1in}
\paragraph{Actor critic model}
Given the encoded information from the sensors over time, the agent needs to decide what action to execute in the environment during exploration. We learn the exploration policy via the actor-critic paradigm, which consists of an \emph{actor} to sample actions and a \emph{critic} to model the value of the agent's current state~\cite{sutton2018reinforcement}. The actor and critic are each modeled using two fully connected layers. The actor outputs the action probabilities for the policy at each time-step, and the critic predicts a scalar state value.

This policy architecture facilitates long-term information storage and effective planning by exploiting geometric and semantic cues in the environment. Unlike traditional SLAM, such a learned spatio-temporal memory---popular in recent embodied perception approaches~\cite{gupta2017cognitive,henriques2018mapnet,chen2019learning,chaplotmodular}---allows the agent to leverage both statistical visual patterns learned from training environments as well as geometry to extrapolate what it learns to novel environments.

\vspace{-0.1in}
\subsection{Policy learning algorithm}\label{sec:policy_learning_algo}
In order to train the policy architecture, we adapt current methods~\cite{embodiedqa,das2018neural,chen2019learning} and divide the learning into two stages: (1) imitation learning, and (2) fine-tuning with reinforcement learning. The imitation learning stage is intended to reduce sample complexity for the reinforcement learning stage by offering a better initialization of the policy.

\vspace{-0.1in}
\paragraph{Imitation learning:} We first pre-train the exploration policy via imitation learning~\cite{bojarski2016end,giusti2016machine}. We generate synthetic expert trajectories by treating the environment as a graph of connected nodes and exploiting this connectivity graph to greedily explore the environment (discussed more in~Sec.~\ref{sec:baselines}). Given these trajectories, we minimize the cross-entropy error between the predicted and expert actions at each step. Additionally, we diversify the training trajectories by sampling actions from the agent's policy with probability $\epsilon$, instead of exclusively following the expert trajectories. We use inflection weighting to prevent the policy from simply repeating the previous action~\cite{wijmans2019embodied}. The perturbation $\epsilon$ is varied from $\epsilon_{start}$ to $\epsilon_{end}$ during the course of training with a constant increase of $0.1$ after every $E$ updates. For AVD, $\epsilon_{start}=0$, $\epsilon_{end}=0.3$, $E=100$. For MP3D, $\epsilon_{start}=0$, $\epsilon_{end}=0.5$, $E=1000$. These values were obtained through validation on the validation environments.

\vspace{-0.1in}
\paragraph{Reinforcement learning:} We then fine-tune the policy with the RL training objective using Proximal Policy Optimization (PPO)~\cite{schulman2017proximal}. The PPO surrogate objective is: 
\begin{equation}
    \label{eqn:ppo_loss}
    L^{CLIP}(\theta) = \hat{\mathbb{E}}_{t}\bigg[ \textrm{min}(\bar{r}_t(\theta)\hat{A}_t,~\textrm{clip}(\bar{r}_t(\theta), 1-\epsilon, 1+\epsilon)\hat{A}_t\bigg],
\end{equation}
where $\theta$ are the policy parameters, $\bar{r}_t(\theta) = \frac{\pi_\theta(a_t|s_t)}{\pi_{\theta_{old}}(a_t|s_t)}$  is the probability ratio of the updated and old policies at time $t$, $\hat{A}_t$ is the advantage estimate, and $\epsilon$ is the clipping factor. We optimize for this objective using the Adam optimizer~\cite{kingma2014adam}. More implementation details can be found in Appendix~\ref{appsec:hyperparameters_exploration}.

Given this empirical study framework consisting of standardized task definitions, simulators, policy architecture, and learning algorithm, we can plug in different reward functions to train different exploration agents, as we describe next.

\section{Taxonomy of exploration paradigms}

\label{sec:taxonomy}
We now present a taxonomy for exploration algorithms in the literature. We identify four core paradigms (see Fig.~\ref{fig:paradigms}): \\
(1) \texttt{\small curiosity}: encourages visiting states where the agent's uncertainty is high.\\
(2) \texttt{\small novelty:} encourages visiting states that are unvisited.\\
(3) \texttt{\small coverage:} encourages visiting states that reveal unseen areas in the environment.\\
(4) \texttt{\small reconstruction:} encourages visiting states that lead to better prediction of unseen states.\\
\indent Each paradigm can be viewed as a particular reward function in the POMDP. In the following, we review their key ideas and related work, and choose representative methods for benchmarking that capture the essence of each paradigm.

\subsection{Curiosity}
\begin{figure*}[t]
\centering
    \includegraphics[width=0.9\textwidth]{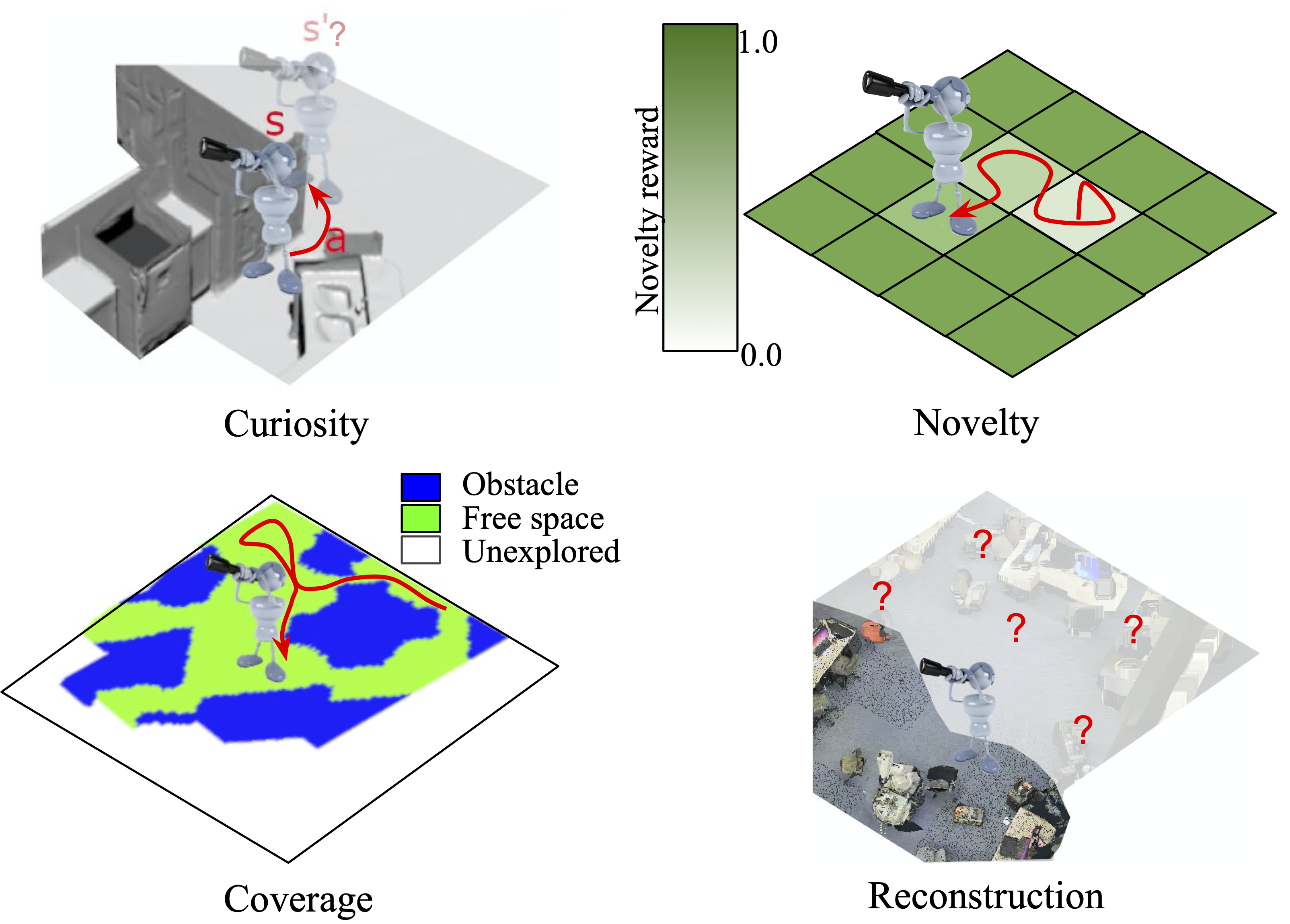}
    \caption{The four paradigms of exploration in 3D visual environments. 
    \textbf{Curiosity} rewards visiting states that are predicted poorly by the current forward-dynamics model.  
    \textbf{Novelty} rewards visiting less frequently visited states. 
    \textbf{Coverage} rewards visiting ``all possible" parts of the environment. 
    \textbf{Reconstruction} rewards visiting states that allow better reconstruction (hallucination) of the full environment.}
    \label{fig:paradigms}
\end{figure*}

In the curiosity paradigm, the agent is encouraged to visit states where its predictive model of the environment is uncertain~\cite{schmidhuber1991curious,oudeyer2007intrinsic,lopes2012exploration,pathak2017curiosity}. See Fig.~\ref{fig:paradigms}, top left.  We focus on the dynamics-based formulation of curiosity, which was shown to perform well on large-scale scenarios~\cite{pathak2017curiosity,burada2018curiosity}. The agent learns a forward-dynamics model $\mathcal{F}$ that predicts the effect of an action on the agent's state, i.e, $\hat{s}_{t+1} = \mathcal{F}(s_{t}, a_{t})$. Then, the \texttt{\small curiosity} reward at each time step is:

\begin{equation}
    r_{t} \propto ||\hat{s}_{t+1} - s_{t+1}||_{2}^{2}.
\end{equation}

The forward-dynamics model is trained online to minimize the loss in forward-dynamics prediction: $||f(s_{t}, a_{t}) - s_{t+1}||_{2}^{2}$. Thus, the \texttt{\small curiosity} reward encourages the agent to move to newer states once its forward-dynamics predictions on the past states are accurate. We adapt the curiosity formulation from~\cite{pathak2017curiosity} for our experiments. Instead of using image features as the state representation for the forward dynamics model, we use the GRU hidden state in Sec.~\ref{sec:policy} as we found it helped alleviate the partial observability. We follow recommended practices such as using PPO for policy optimization, reward normalization, observation normalization, and using feature normalization in the CNN~\cite{burada2018curiosity}.

\paragraph{Past work on curiosity-driven exploration} 
The forward-dynamics based formulation of curiosity uses an inverse-dynamics prediction loss to learn features that ignore parts of the environment not useful for dynamics prediction~\cite{pathak2017curiosity}. Stable features kept fixed during policy learning are found to perform better in many cases~\cite{burada2018curiosity}. Further variants of curiosity aim to overcome the ``noisy-TV" problem where the curiosity agent fixates on parts of the environment where stochasticity is a function of its actions~\cite{burada2018curiosity,burda2018exploration,pathak2019self}. For example, if the agent has access to a TV remote that allows it to randomly change channels, it does that repeatedly to accumulate rewards. Due to the inherently random nature of the transitions, they can never be predicted accurately and always reward the agent. Random Network Distillation attempts to overcome this limitation by defining a fixed random neural network as the predictive function since it is a deterministic function of the inputs~\cite{burda2018exploration}. Exploration via Disagreement attempts to resolve the issue by learning an ensemble of forward-prediction models and using their disagreement as intrinsic motivation rather than the prediction error~\cite{pathak2019self}. Curiosity is also demonstrated to work on physically realistic 3D simulators with the addition of a self-aware model that learns to anticipate errors in the agent's predictions~\cite{haber2018selfaware}.

\subsection{Novelty}
While curiosity seeks hard-to-predict states, novelty seeks previously unvisited states~\cite{strehl2008analysis,bellemare2016unifying,ostrovski2017count,tang2017exploration,savinov2018episodic}.  See Fig.~\ref{fig:paradigms}, top right. Each $s \in \mathcal{S}$ is assigned a visitation count $n(s)$. The \texttt{\small novelty} reward is inversely proportional to the square-root of visitation frequency at the current state: 

\begin{equation}
   r_{t} \propto \frac{1}{\sqrt{n(s_{t})}}.
   \label{eqn:novelty}
\end{equation}
For our experiments, we adapt the Grid Oracle method from~\cite{savinov2018episodic}: we discretize the 3D environment into a 2D grid where each grid cell is considered to be a unique state, and assign rewards according to Eqn.~\ref{eqn:novelty}. We define square grid cells of width $0.3\si{m}$ in AVD and $0.5\si{m}$ in MP3D, and consider all points within a grid cell to correspond to that state.  

\paragraph{Past work on novelty-based exploration}
Early methods on interval estimation or count-based exploration define a confidence interval over the optimal value function of the underlying Markov Decision Processes (MDP) and apply this idea to tabular RL settings~\cite{strehl2008analysis}. More recent work extends this idea to function approximation settings by using the notion of pseudo-counts~\cite{bellemare2016unifying} combined with complex density models~\cite{ostrovski2017count}, or discretizing high-dimensional continuous spaces into discrete representations through hash functions~\cite{tang2017exploration}, which facilitates a tabular RL setting. Rather than using dynamics prediction, episodic curiosity~\cite{savinov2018episodic} rewards states that look dissimilar from previously visited ones, yielding a binary variant of Eqn.~\ref{eqn:novelty}. We use the GridOracle from~\cite{savinov2018episodic} as it was shown to explore successfully in a variety of settings.

\subsection{Coverage} 
\label{sec:coverage}
The coverage paradigm aims to observe as many things of interest as possible---typically the area seen in the environment~\cite{chen2019learning,chaplotmodular}. See Fig.~\ref{fig:paradigms}, bottom left. Whereas novelty encourages explicitly visiting all locations, coverage encourages \emph{observing} all of the environment. Note that the two are distinct: at any given location, how much and how far a robot can see varies depending on the surrounding 3D structures. The coverage approach from~\cite{chen2019learning} learns RL policies that maximize area coverage. The \texttt{\small area-coverage} reward is:
\begin{equation}
    \label{eqn:coverage}
    r_{t} \propto A_{t} - A_{t-1},
\end{equation}
\noindent where $A_{t}$ is the area covered by the agent till time $t$ (blue and green regions in Fig.~\ref{fig:policy_architecture}). An agent maximizing area seen prioritizes those locations and viewing angles which provide high visibility of unseen parts of the environment. 

While this is shown to work well on diverse environments~\cite{chen2019learning}, the agent is never rewarded for revisiting the same regions. In our experiments, we find that this leads to a specific form of reward sparsity in large environments (as we will highlight in Sec.~\ref{sec:ablation_studies}). Consider the case where an agent starts exploring from the center of a very large building. After it explores one half of the building, there will be no learning signal for the agent to move back to the center in order to continue exploring the other half. Therefore, we incorporate ideas from the \texttt{\small novelty} reward to design a shaped version of \texttt{\small coverage} that addresses this limitation, which we call \texttt{\small smooth-coverage}:
\begin{equation}
    r_{t} \propto \frac{1}{|\text{\textbf{A}}_{t}|} \sum_{i \in \text{\textbf{A}}_t} \frac{1}{\sqrt{n_i}},
\end{equation}
where $\text{\textbf{A}}_t$ is the set of regions seen in the environment at time $t$ and $n_i$ is the number of times a region $i$ has been observed so far. Instead of only rewarding the agent for the first time it sees a region (as in Eq.~\ref{eqn:coverage}), we reward the agent based on the number of times that region was observed. This allows the agent to navigate across less frequently visited regions to discover unexplored parts of the environment.

While we use area as our primary quantity of interest, this idea can more generally be exploited for learning to visit other interesting things such as objects (similar to the search task from~\cite{fang2019scene}) and landmarks. See Appendix~\ref{appsec:comparing_coverage_variants} for comparative results.  

\paragraph{Past work on coverage-based exploration}
Recent work explores using novel architectures that better optimize the coverage objective. The area coverage objective is optimized through a combination of imitation and reinforcement learning to learn exploration policies that generalize well to novel environments~\cite{chen2019learning}. A new hierarchical architecture combines classic-planning and learned controllers to successfully optimize the area coverage objective in a sample-efficient manner under realistic noise models~\cite{chaplotmodular}. Scene memory transformers propose an architecture based on transformers~\cite{vaswani2017attention} that encode observations at each time-step into a feature space where self-attention is used to perform reasoning~\cite{fang2019scene}. The model is shown to be successful on several tasks including area coverage and object class coverage (i.e., visiting as many object classes as possible). These architectural variants are orthogonal to our study since our primary goal is to explore the choice of reward function to facilitate exploration.


\subsection{Reconstruction} 
\label{sec:reconstruction}
Reconstruction-based methods~\cite{dinesh2018ltla,ramakrishnan2018sidekick,ramakrishnan2019emergence,seifi2019look} use the objective of active observation completion~\cite{dinesh2018ltla} to learn exploration policies. The idea is to gather views that best facilitate the prediction of unseen viewpoints in the environment. See Fig.~\ref{fig:paradigms}, bottom right.  The \texttt{\small{reconstruction}} reward scores the quality of the predicted outputs:
\begin{equation}
    \label{eq:recons}
    r_{t} \propto -d(V(\mathcal{P}), \hat{V}_{t}(\mathcal{P})),
\end{equation}
where $V(\mathcal{P})$ is a set of true ``query" views at camera poses $\mathcal{P}$ in the environment, $\hat{V}_{t}$ is the set of view reconstructions generated by the agent after $t$ time steps, and $d$ is a distance function. Whereas \texttt{\small curiosity} rewards views that are individually surprising, \texttt{\small reconstruction} rewards views that bolster the agent's correct hallucination of all other views. 

Prior attempts at exploration with reconstruction are limited to pixelwise reconstructions on $360^{\circ}$ panoramas and CAD models, where $d(\cdot,\cdot)$ is $\ell_2$ on pixels. To scale the idea to 3D environments, we propose a novel adaptation that predicts \emph{concepts} present in unobserved views rather than pixels, i.e., a form of semantic reconstruction. This reformulation requires the agent to predict whether, say, an instance of a ``door'' concept is present at some query location, rather than reconstruct the door pixelwise as in~\cite{dinesh2018ltla,ramakrishnan2019emergence}. 

We automatically discover these visual concepts from the training environments, which has the advantage of not relying on supervised object detectors or semantic annotations. Specifically, we sample views uniformly from training environments and cluster them into discrete concepts $\mathcal{C} = \{c_{i}\}_{i=1}^{K}$ using $K$-means applied to ResNet-50 features.  Each cluster centroid is a concept and may be semantic (doors, pillars) or geometric (arches, corners). 

Then, we reward the agent for acquiring views that help it accurately predict the dominant concepts in all query views $V(\mathcal{P})$ sampled from a uniform grid of locations in each environment. Let $v(p_{i}) \in V(\mathcal{P})$ denote the ResNet-50 feature for the $i$-th query view. We define its true ``reconstructing" concepts $C_i = \{c_{i}^{1},\dots,c_{i}^{J}\} \subset \mathcal{C}$ to be the $J$ nearest cluster centroids to $v(p_{i})$, and assign equal probability to those $J$ concepts. The agent has a multilabel classifier that takes a query pose $p_{i}$ as input---\emph{not} the view $v(p_{i})$---and returns $\hat{C}_i$, the posteriors for each concept $c \in \mathcal{C}$ being present in $v(p_{i})$. The distance $d$ in Eqn~\eqref{eq:recons} is the KL-divergence between the true concept distribution $C_i$ and the agent's inferred distribution $\hat{C}_i$, summed over all $p_{i} \in \mathcal{P}$. The reward $r_{t}$ thus encourages reducing the prediction error in reconstructing the true concepts. In practice, we shape the reward function to be the reduction in the KL divergence over time (as opposed to the absolute KL divergence value) as this leads to better performance; see Appendix~\ref{appsec:reconstruction_pipeline}.

\paragraph{Past work on reconstruction-based exploration}
The original reconstruction-based exploration approach~\cite{dinesh2018ltla} rewards exploration that permits more accurate pixelwise reconstructions of scenes and objects. It was shown that the resulting exploratory policies accelerate recognition tasks. Sidekick policy learning~\cite{ramakrishnan2018sidekick} exploits full state information available \textit{only} at training time to accelerate training and learn better policies. Broader transferability of exploration policies on the tasks of recognition, volume estimation, and other tasks are demonstrated in~\cite{ramakrishnan2019emergence}, and the quality of pixelwise reconstructions are enhanced using a pix2pix~\cite{pix2pix2016} model.  Architectural improvements based on a spatial memory and foveated viewpoints yield high-quality reconstructions with lower viewing costs~\cite{seifi2019look}.

Reconstruction-based exploration approaches also relate to classical information-gain based exploration, where the goal is to explore in order to reduce the uncertainty in the agent's model of the world~\cite{cassandra1996acting,stachniss2005information,sun2011planning}. The action entropy control strategy in~\cite{cassandra1996acting} explores in order to maximize the information gained about the agent's belief state when the certainty of the optimal action is low. In~\cite{stachniss2005information}, the agent is encouraged to take actions that minimize the uncertainty in pose estimation and mapping. Optimal Bayesian exploration is used to maximize the agent's knowledge about a parameterized world model~\cite{sun2011planning}. Our proposed formulation that reconstructs in a discrete concept space can be viewed as a form of information-gain based exploration. The high-dimensional image-state representation is projected to a smaller concept space, and the reward function encourages maximizing the the information gain between the agent's knowledge of the concepts present in the environment and the states visited by the agent.

\section{Exploration evaluation framework}
\label{sec:evaluation_framework}
Having defined the taxonomy of exploration algorithms, we now define baselines and metrics to evaluate them.

\subsection{Baseline methods}\label{sec:baselines}
\noindent\textbf{Heuristic baselines:} Aside from the learned paradigms introduced above, we also evaluate four non-learned heuristics for exploration:\\
(1) \texttt{\small random-actions}~\cite{ramakrishnan2019emergence,habitat19iccv} samples from a uniform distribution over all actions. \\
(2) \texttt{\small forward-action}~\cite{habitat19iccv} always samples the forward action. \\
(3) \texttt{\small forward-action+} samples the forward action unless a collision occurs, in which case, it turns left. \\ 
(4) \texttt{\small frontier-exploration}~\cite{yamauchi1997frontier} uses the egocentric map from Fig.~\ref{fig:policy_architecture} and iteratively visits the frontiers, i.e., the edges between free and unexplored spaces (see Appendix~\ref{appsec:frontier_exploration}). The \texttt{\small frontier} approach is closely related to \texttt{\small area-coverage}, but depends on hand-crafted heuristics and may be vulnerable to noisy sensors and actuators.

\noindent\textbf{Oracle graph exploration:}  As an upper bound on exploration performance, we also include an oracle model. It cheats by exploiting the underlying graph structure in the environment (which reveals reachability and obstacles) to visit a sequence of sampled target locations via the true shortest paths. In contrast, all methods we benchmark are \emph{not} given this graph, and must discover it through exploration. We define three \texttt{\small oracles} that visit (1) randomly sampled locations, (2) landmark views (see Sec.~\ref{sec:evaluation_metrics}), and (3) objects within the environment.

\noindent\textbf{Imitation learning:} We imitate the \texttt{\small oracle} trajectories to get three \texttt{\small imitation} variants, one for each oracle above. Whereas the oracles assume full observability and therefore are not viable exploration approaches, these imitation learning baselines are viable, assuming disjoint training data with full observability is available.

Tab.~\ref{tab:assumptions} lists the assumptions on information availability made by the different approaches. Methods requiring less information are more flexible. While we assume access to the full environment during training, there is ongoing work~\cite{bellemare2016unifying,ostrovski2017count,savinov2018episodic} on relaxing this assumption.

\begin{table}[t]
\centering
\resizebox{0.8\textwidth}{!}{
\begin{tabular}{@{}lcccc@{}}
\toprule
\multicolumn{1}{c}{}                   & \multicolumn{4}{c}{Training / Testing} \\ 
\cmidrule{2-5}  
\multicolumn{1}{c}{\multirow{-2}{*}{}} & GT depth                                               &                     GT pose                                  & GT objects                                             & GT state                      \\ \midrule
\texttt{ random-actions}          &       ~~~~    -               / {\color[HTML]{32CB00} No}   &    ~~~~     -                 / {\color[HTML]{32CB00} No}    &      ~~~~-                / {\color[HTML]{32CB00} No}  &           ~~~~-             / {\color[HTML]{32CB00} No}  \\
\texttt{ forward-action(+)}       &       ~~~~    -               / {\color[HTML]{32CB00} No}   &     ~~~~    -                 / {\color[HTML]{32CB00} No}    &    ~~~~  -                / {\color[HTML]{32CB00} No}  &          ~~~~ -             / {\color[HTML]{32CB00} No}  \\
\texttt{ imitation-X}             & {\color[HTML]{FE0000}~~Yes  } / {\color[HTML]{F8A102} Yes*} & {\color[HTML]{FE0000}~~Yes  } / {\color[HTML]{F8A102} Yes*}  & {\color[HTML]{FE0000}~Yes} / {\color[HTML]{32CB00} No}  & {\color[HTML]{FE0000}~Yes  }/ {\color[HTML]{32CB00} No}  \\
\texttt{ curiosity}               & {\color[HTML]{FE0000}~~Yes  } / {\color[HTML]{F8A102} Yes*} & {\color[HTML]{FE0000}~~Yes  } / {\color[HTML]{F8A102} Yes*}  & {\color[HTML]{32CB00} No } / {\color[HTML]{32CB00} No}  & {\color[HTML]{32CB00} No }  / {\color[HTML]{32CB00} No}  \\
\texttt{ novelty}                 & {\color[HTML]{FE0000}~~Yes  } / {\color[HTML]{F8A102} Yes*} & {\color[HTML]{FE0000}~~Yes  } / {\color[HTML]{F8A102} Yes*}  & {\color[HTML]{32CB00} No } / {\color[HTML]{32CB00} No}  & {\color[HTML]{FE0000}~Yes  }/ {\color[HTML]{32CB00} No}  \\
\texttt{ frontier-exploration}    &                        ~~~~~- / {\color[HTML]{FE0000} Yes } &                      ~~~~~-   / {\color[HTML]{FE0000} Yes }  &                     ~~~~~- / {\color[HTML]{32CB00} No}  &                      ~~~~~- / {\color[HTML]{32CB00} No}  \\
\texttt{ coverage}                & {\color[HTML]{FE0000}~~Yes  } / {\color[HTML]{F8A102} Yes*} & {\color[HTML]{FE0000}~~Yes  } / {\color[HTML]{F8A102} Yes*}  & {\color[HTML]{FE0000}Yes } / {\color[HTML]{32CB00} No}  & {\color[HTML]{32CB00} No }  / {\color[HTML]{32CB00} No} \\
\texttt{ oracle}                  & {\color[HTML]{32CB00} No }    / {\color[HTML]{32CB00} No}   & {\color[HTML]{32CB00} No }    / {\color[HTML]{32CB00} No}    & {\color[HTML]{FE0000} Yes} / {\color[HTML]{FE0000} Yes }  & {\color[HTML]{FE0000}~Yes  } / {\color[HTML]{FE0000} Yes }  \\
\bottomrule
\vspace{0.1cm}
\end{tabular}
}
\caption{\textbf{Assumptions about information availability}: the information required for each method (including the architecture assumptions) during training/testing. In our experiments, we assume all information is given during training, but only sensory inputs are given for testing. * - learned methods may adapt to noisy inputs.
}

\label{tab:assumptions}
\end{table}

\subsection{Evaluation metrics}
\label{sec:evaluation_metrics}
A good exploration method visits interesting locations and collects information that is useful for a variety of downstream tasks. Different methods may be better suited for different tasks. For example, a method optimizing for area coverage may not interact sufficiently with objects, leading to poor performance on object-centric tasks. We measure exploration performance with two families of metrics:

\paragraph{(1) Visiting interesting things.} 
These metrics quantify the extent to which the agent visits things of interest such as area~\cite{chen2019learning,yamauchi1997frontier,chaplotmodular,fang2019scene}, objects~\cite{fang2019scene,haber2018selfaware}, and landmarks. The area visited metric measures the amount of area covered in $\si{m^2}$. For the objects visited metric, we use the object annotations provided by both AVD and MP3D to measure how many objects are visited during exploration. We consider an object to be visited if the agent is close to it and the object is unoccluded within its field of view. For the landmarks visited metric, we mine a set of ``landmark'' viewpoints from the environment which contain distinctive visual components that do not appear elsewhere in that environment, e.g., a colorful painting or a decorated fireplace.\footnote{See Appendix~\ref{appsec:mining_landmarks} for more details on how landmarks are mined.} The exact visitation criteria for both objects and landmarks are given in Appendix~\ref{appsec:visitation_criteria}. Together, these visitation metrics capture the different levels of semantic and geometric reasoning that an agent may need to perform in the environment. To account for varying environment sizes and content, we normalize each metric into $[0,1]$ by dividing by the best \texttt{\small oracle} score on each episode.

\paragraph{(2) Downstream task transfer.}
These metrics directly evaluate exploration's impact on downstream tasks. The setup is as follows: an exploration agent is given a time budget of $T_{exp}$ to explore the environment, after which the information gathered must be utilized to solve a task within the same environment. More efficient exploration algorithms gather better information and are expected to have higher task performance. 

We consider three tasks from the literature that ask fundamental, yet diverse questions: (i) \emph{PointNav}: how to quickly navigate from point A to point B? (ii) \emph{view localization}: where was this photo taken? and (iii) \emph{reconstruction}: what can I expect to see at point B?

In \emph{PointNav}, the agent is respawned at the original starting point after exploration, and given a navigation goal $(x, y, z)$ relative to its position~\cite{anderson2018evaluation,savva2017minos}. The agent must use its map acquired during exploration to navigate to the goal within a maximum of $T_{nav}$ time steps. Intuitively, for efficient navigation, an exploration algorithm needs to explore potential dead ends in the environment that may cause planning failure. We use an A* planner that navigates using the spatial occupancy map built during exploration~\cite{hart1968formal}. While other navigation algorithms are possible, A* is a consistent and lightweight means to isolate the impact of the exploration models. We evaluate using the success rate normalized by the path length (SPL)~\cite{anderson2018evaluation}.

In \emph{view localization}, the agent is presented with images sampled from distinct landmark viewpoints (introduced previously in Sec.~\ref{sec:evaluation_metrics}) and must localize them relative to its starting location~\cite{ramakrishnan2019emergence}. This task captures the agent's model of the overall layout, e.g., ``where would you need to go to see this view?". While PointNav requires \emph{planning} a path to a \emph{point} target around obstacles, view localization requires \emph{locating} a \emph{visual} target. We measure localization accuracy by the pose-success rate (PSR$@ k$), the fraction of views localized within $k$ mm of their ground truth pose.

In \emph{reconstruction}, the agent is presented with uniformly spread query locations, and must predict the set of concepts present at each location~\cite{dinesh2018ltla,ramakrishnan2019emergence,seifi2019look} (cf.~Sec.~\ref{sec:reconstruction}). This can be viewed as the inverse problem of view localization: the agent has to predict views given locations. Performance is measured using Precision$@K$ between the agent's predicted concepts and the ground truth. 

For each task, we define a standard pipeline that uses the exploration experience to solve the task in Appendix A. We do not fine-tune exploration policies for downstream tasks since we treat them as independent evaluation metrics. 

Together, the proposed metrics capture both geometric and semantic aspects of spatial sensing.  For example, the area-visited metric reflects 2D SLAM (occupancy) and  the reconstruction task metric measures semantic SLAM, where the goal is to map concepts present in the environment.
\section{Experiments}
\label{sec:experiments}

We next present the results on various exploration metrics.

\paragraph{Implementation details}
All policies are trained for $T_{exp}=200/500$ on AVD/MP3D for 2000 episode batches. We sample $100/342$ episodes on AVD/MP3D uniformly from all test environments. For navigation, we define $46/195$ difficult test episodes by selecting challenging pairs of start-goal positions on AVD/MP3D. They are selected such that the agent's navigation would be inefficient unless the environment was already well explored prior to navigation (see Appendix~\ref{appsec:difficult_pointnav_episodes}). More implementation details are provided in Appendix~\ref{appsec:hyperparameters_exploration}. 

\subsection{Results on exploration metrics}
\label{sec:exploration_results}
We compare the performance of different paradigms on visitation metrics and downstream task transfer in AVD and MP3D. Since the two datasets exhibit a variety of contrasting properties by design (see Sec.~\ref{sec:datasets}), we may expect to see corresponding differences in the results. For brevity, only the best of the three \texttt{\small oracle} and \texttt{\small imitation} variants on each metric are reported. Note that \texttt{\small coverage} here corresponds to \texttt{\small smooth-coverage}, which we found to consistently outperform \texttt{\small area-coverage}.

\begin{figure}[!]
\centering
\includegraphics[width=1.0\textwidth,trim={0.2cm 0 0.2cm 0},clip]{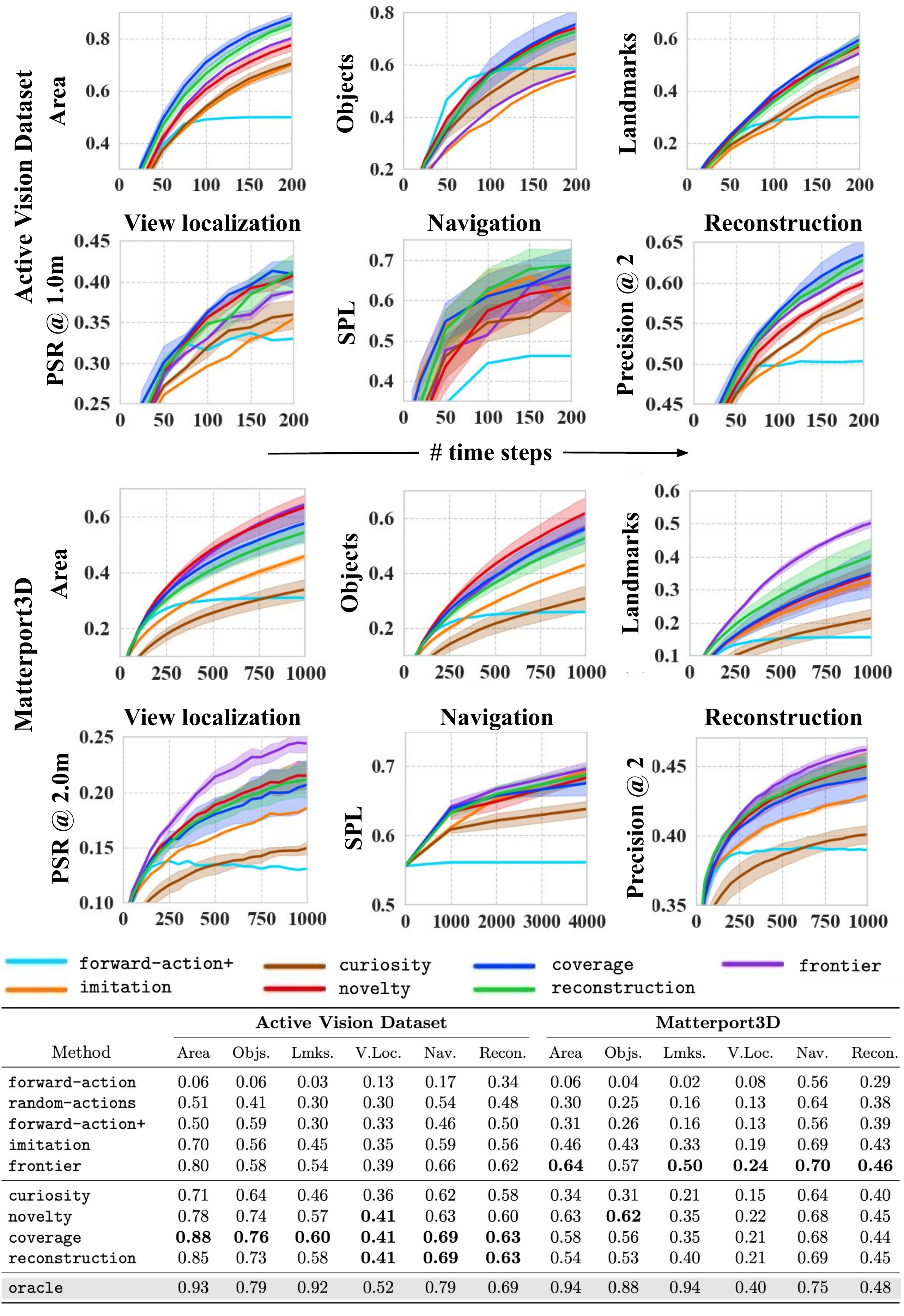}
\caption{\textbf{Exploration results:} The plots compare the different paradigms and select baselines (for clarity) on all exploration metrics. The table shows the performance averaged over three random seeds on each metric at the last time step and includes all baselines. The visitation metrics are area-, object-, and landmark-visitation. The downstream tasks are view localization, navigation, and reconstruction. The best results are bolded.}
\label{fig:coverage_results}
\end{figure}

\begin{figure}[!]
    \centering
    \includegraphics[width=1.0\textwidth]{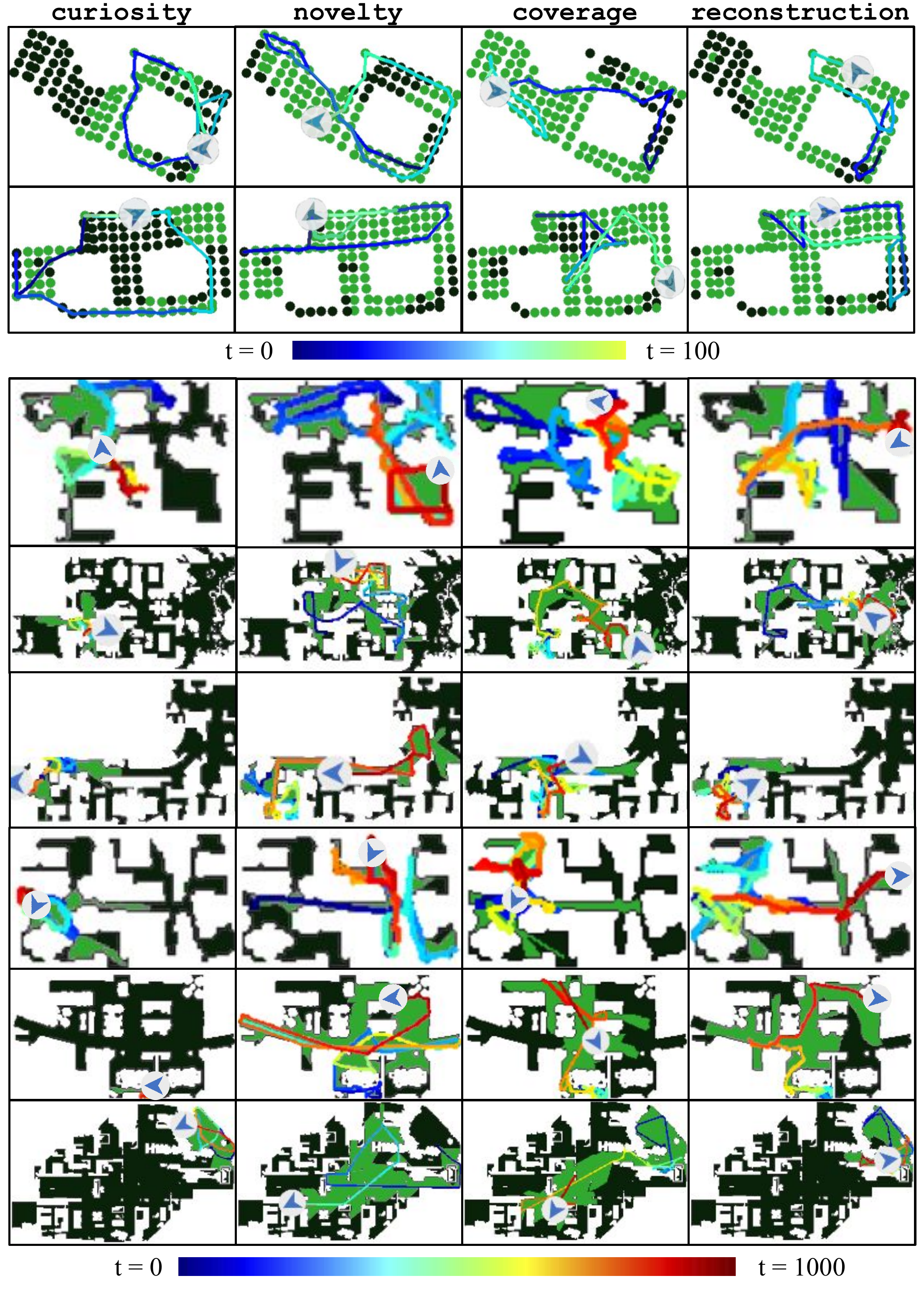}
    \caption{\textbf{Visualizing exploration behaviors:} Exploration trajectories for each paradigm are visualized from the top-down view of the environment (AVD in first two rows, MP3D in the remaining rows). Black and green locations represent unexplored and explored areas, respectively. The agent's trajectory uses color changes to represent time. The behaviors are largely correlated with the quantitative performance of each paradigm in Sec.~\ref{sec:exploration_results}: better exploration methods cover larger parts of the environment. For example, in AVD, \texttt{coverage} explores efficiently and observes most of the environment within 100 time-steps, closely followed by \texttt{reconstruction} and \texttt{novelty} (top two rows). In MP3D, \texttt{novelty} covers significantly larger areas during exploration in large environments (rows 5, 7 and 8). In fact, \texttt{novelty} is the only approach that successfully crosses the narrow corridor in the middle and explores both sides of the environment in row 5.}
    \label{fig:qual_exploration}
\end{figure}

Fig.~\ref{fig:coverage_results} shows the results on both datasets, and Fig.~\ref{fig:qual_exploration} shows example exploration episodes. These results are for the ideal noise-free odometry case. We will separately analyze the impact of noisy odometry in the next section. We see two clear trends emerging:
\begin{align}
    \textrm{\textbf{AVD:}}~~~\textrm{coverage} > \textrm{reconstruction} > \textrm{novelty} > \textrm{curiosity}\nonumber \\
    \textrm{\textbf{MP3D:}}~~\textrm{novelty} > \textrm{coverage} \ge \textrm{reconstruction} > \textrm{curiosity}\nonumber
\end{align}
In AVD, both \texttt{\small coverage} and \texttt{\small reconstruction} clearly outperform \texttt{\small novelty}. While \texttt{\small novelty} assigns equal rewards to all locations in the environment, \texttt{\small coverage} and \texttt{\small reconstruction} prioritize viewpoints that are useful for quickly covering area and concepts, respectively. This significantly improves exploration performance in the small, cluttered rooms of AVD. However, we do not observe a similar effect in MP3D as \texttt{\small novelty} outperforms both these methods. This is due to two reasons. First, since the rooms in MP3D are generally open and large, the advantage of prioritizing the viewpoints is smaller than in AVD. Second, the reward functions of \texttt{\small coverage} and \texttt{\small reconstruction} are inherently sparser than \texttt{\small novelty}. The \texttt{\small reconstruction} method does not reward the agent for reconstructing concepts that it already reconstructed in the past (reduction in KL divergence is smaller). The original \texttt{\small area-coverage} method does not reward the agent for seeing the same area more than once. As we will show in an ablation in Sec.~\ref{sec:ablation_studies}, the improved \texttt{\small smooth-coverage} method bridges the gap between \texttt{\small area-coverage} and \texttt{\small novelty} to a good extent.

The performance of \texttt{\small curiosity} is worse than that of other paradigms on both datasets, likely due to the significant partial observability that is characteristic of these visually rich 3D environments. Its performance deteriorates further on MP3D environments which tend to be much larger, and it even underperforms the \texttt{\small imitation} baseline. \texttt{\small curiosity} needs a good state representation that accounts for partial observability to learn a good forward dynamics model that leads to better reward estimates. This is naturally harder in larger MP3D environments. As we will show in ablations in Sec.~\ref{sec:ablation_studies}, the memory architecture that we use here is better for \texttt{\small curiosity} than using only image features.  It is possible that incorporating better memory architectures could boost the performance even further. 

Good baselines are essential for measuring progress on any benchmark. Our proposed baseline \texttt{\small imitation} is significantly better than the standard baselines \texttt{\small random-actions}, \texttt{\small forward-action}, and \texttt{\small forward-action+} used in prior work~\cite{ramakrishnan2019emergence,habitat19iccv,chen2019learning}. While learned methods generally outperform these baselines, \texttt{\small frontier-exploration} outperforms most learned methods on MP3D. Since \texttt{\small fronter-exploration} relies on clean depth inputs with a perfectly registered map, it underperforms in AVD since the depth inputs are naturally noisy (see Sec.~\ref{fig:3d_environments}). As we will show in the next section, a similar phenomenon can also be in observed in MP3D where noisy sensor inputs deteriorate its performance, echoing past findings~\cite{chen2019learning}. 

In Fig.~\ref{fig:coverage_results}, we see that the navigation performance for all methods in MP3D are inferior to the state-of-the-art performance reported in standard navigation benchmarks on MP3D~\cite{habitat19iccv} due to the increased difficulty of the test episodes. Additionally, we observe that the trends on the navigation metric on both AVD and MP3D differ from the other metrics. The gaps between different methods are reduced, \texttt{\small imitation} closely competes with other paradigms, and much larger values of $T_{exp}$ are required to improve the performance in MP3D. This hints at the inherent difficulty of exploring well to navigate: efficient navigation requires exploration agents to uncover potential obstacles that reduce path planning efficiency. Notably, existing exploration methods do not incorporate such priors into their rewards. 

\begin{table}[t!]
    \begin{minipage}{\linewidth}
        \centering
        \includegraphics[width=0.95\textwidth,trim={0 0 0 0},clip]{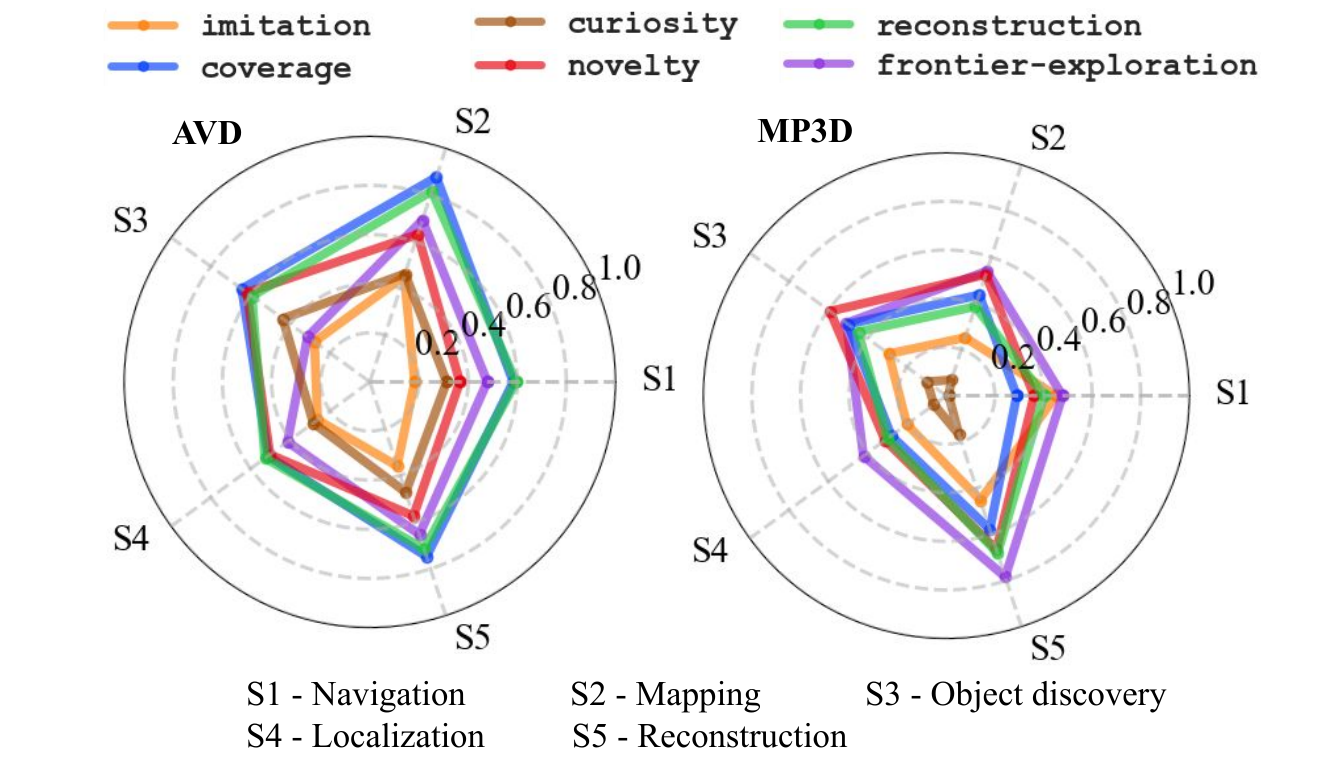}
        \captionof{figure}{\textbf{Exploration skills:} Each exploration agent is assigned a 0-1 value for a set of five skills. The most general agents have larger polygon areas (see text). Best viewed in color.}
        \label{fig:radar_plots}
    \end{minipage}
    \vspace{0.4cm}
\end{table}

The radar plots in Fig.~\ref{fig:radar_plots} concisely summarize the results thus far along five skills: Mapping, Navigation, Object discovery, Localization, and Reconstruction. The metrics corresponding to these skills are area visited, pointnav SPL, objects visited, view localization PSR, and reconstruction precision@2, respectively. The metrics are normalized to [0, 1] where 0 and 1 represent the performance of \texttt{\small random-actions} and \texttt{\small oracle}, respectively.  In each case, a method's performance on the metric is normalized as follows:
\begin{equation}
    \textrm{skill value} = \frac{\textrm{method score} - \textrm{random-actions score}}{\textrm{oracle score} - \textrm{random-actions score}}.
\end{equation}

As we can see in Fig.~\ref{fig:radar_plots}, \texttt{\small coverage} dominates the other paradigms on most skills, closely followed by \texttt{\small reconstruction} in AVD. In the larger MP3D environments, different methods are stronger on different skills. For example, \texttt{\small novelty} performs best on object discovery, \texttt{\small frontier-exploration} dominates on localization and reconstruction, and both dominate on mapping. This highlights the need for diverse evaluation metrics for exploration.

Fig.~\ref{fig:qual_exploration} shows qualitative examples of exploration in AVD and MP3D for the four paradigms. Better methods cover larger areas of the environment during exploration.

\subsection{Impact of sensor noise on exploration}
\label{sec:sensor_noise}
In the previous section, we evaluated the exploration performance of various algorithms under noise-free odometry. Now, we evaluate the impact of sensor noise during exploration, specifically the noise arising from two sources: (1) the occupancy map, and (2) odometer readings. We perform our experiments in MP3D as it contains larger and more diverse testing environments. 

\paragraph{Impact of noisy occupancy estimation}
Noisy occupancy maps can result from mesh defects, noisy depth, and incorrect estimates of affordances from the height-based estimation method (see Sec.~\ref{sec:policy})~\cite{qi2020learning}. This noise occurs naturally and is not explicitly simulated, meaning all results reported thus far are where the methods \emph{do} experience noisy occupancy.  The noise-free case can be simulated by directly extracting occupancy maps from the navigability information stored in the simulator, instead of estimating it from visual cues. We characterize the impact of this noise through the \emph{noise robustness coefficient (NRC)}, the ratio of the area visited\footnote{We select area visited because it is a simple metric that does not require additional semantic annotations.} by an agent with the original depth-based occupancy maps (noisy) and ground-truth occupancy maps obtained from the simulator (noise-free). Larger NRC indicates better adaptation to noisy test conditions. 

Tab.~\ref{tab:noise_robustness} shows the exact values of area visited with noisy and noise-free occupancy, and the corresponding NRC values. As we can see, learned approaches have an NRC close to $1.0$, indicating that they perform similarly under both conditions.~\footnote{The NRC values may be larger than $1.0$ for learned methods. This is due to domain differences between the inferred occupancy used in training and the GT occupancy used in testing.} However, a purely geometric approach like \texttt{\small frontier-exploration} has a much lower NRC ($\sim 0.77$). Since the simulation conditions in MP3D are generally noise-free and the noise levels are low, \texttt{\small frontier-exploration} still gets better absolute performance than most approaches. In contrast, the depth maps in AVD lead to persistent noise in the occupancy estimates, which causes \texttt{\small frontier-exploration} to do worse than other approaches there (see Fig.~\ref{fig:coverage_results}).

\begin{table}[t]
\centering
\begin{tabular}{lccc}
\toprule
\multicolumn{4}{c}{Effect of incorrectly estimated occupancy maps on exploration}                       \\ \midrule
Method                                & Inferred occupancy    & GT occupancy         & NRC              \\ \midrule
\texttt{\small imitation}             & 0.452                 & 0.386                & 1.17             \\
\texttt{\small frontier-exploration}  & 0.579                 & 0.748                & 0.77             \\
\texttt{\small coverage}              & 0.522                 & 0.508                & 1.03             \\
\texttt{\small novelty}               & 0.622                 & 0.599                & 1.04             \\
\texttt{\small curiosity}             & 0.317                 & 0.317                & 1.00             \\
\texttt{\small reconstruction}        & 0.510                 & 0.503                & 1.01             \\ \bottomrule
\end{tabular}
\caption{\textbf{Noise robustness:} We show the average area visited by each method with our original depth-based occupancy estimates which tend to be noisy (second column) and the ground-truth occupancy maps obtained from the simulator which are noise-free (third column). The area visited has been normalized by the best oracle score on a per-episode basis. In the fourth column, we show the noise robustness coefficient (NRC) which is the ratio of performance with the original and ground truth (GT) occupancy maps (higher is better).}
\label{tab:noise_robustness}
\end{table}

\paragraph{Impact of noisy odometry}

\begin{figure}[t]
    \centering
    \includegraphics[width=1.0\textwidth]{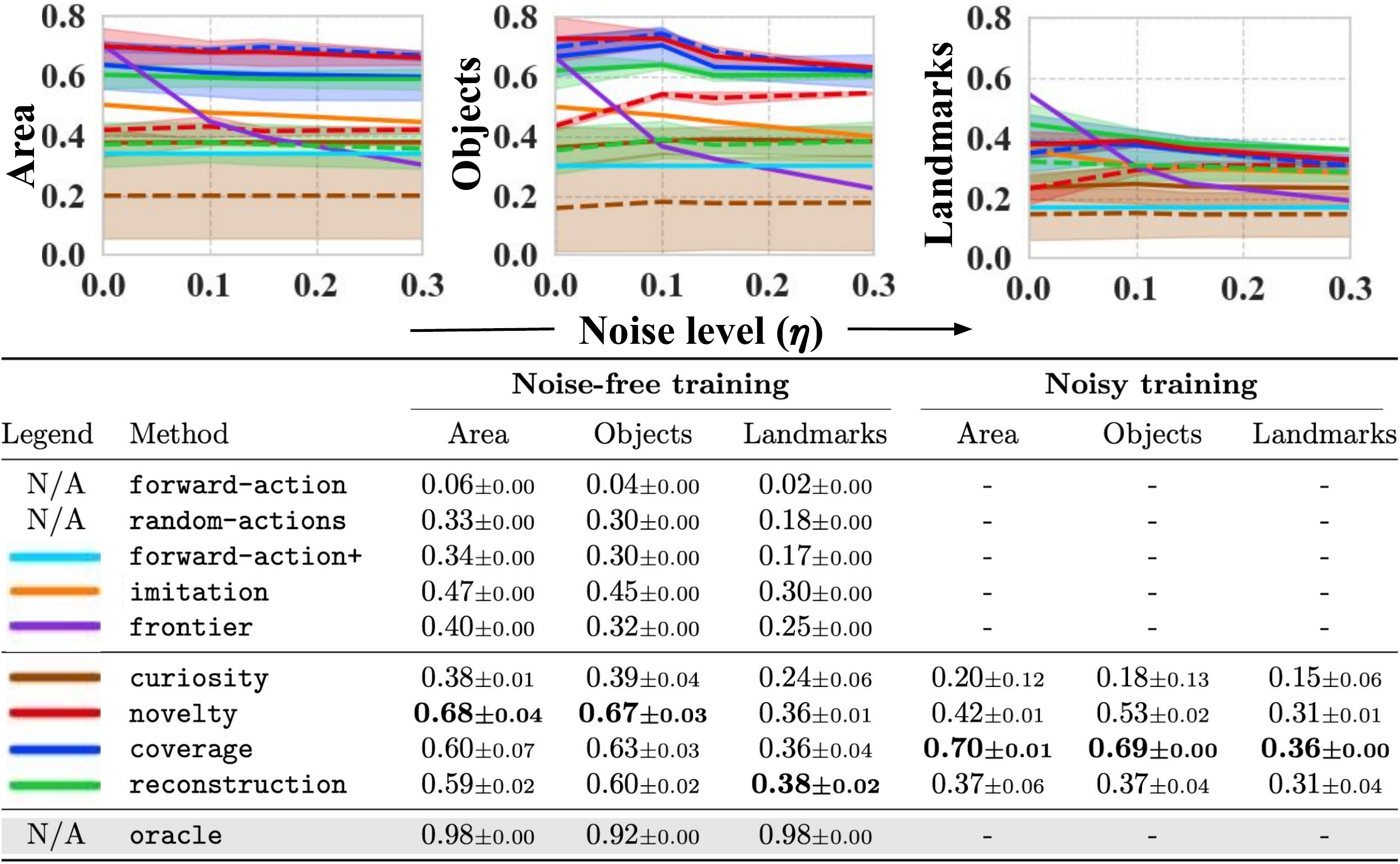}
    \caption{The plots show exploration performance as a function of noise levels in the odometer ($\eta$) during testing.  Note that the legend for the plots is shown in the table. The dashed lines in the plots denote noisy training conditions ($\eta=0.15$). The table compares the performance of various paradigms and baselines with noisy testing conditions ($\eta=0.15$), and both noise-free and noisy training conditions. }
    \label{fig:noisy_odometer}
\end{figure}

Next, we measure the impact of a noisy odometer during the exploration evaluation phase. Noisy odometry leads to incorrect registration of local point clouds, estimated from the depth map to the allocentric map (see Sec.~\ref{sec:policy}). This noise affects the spatial memory of the agent and effectively leads to noisy occupancy inputs to the exploration policy. Note that this is different from the noisy occupancy maps case where we measure the impact of per-frame noise in occupancy. Here, we are more interested in the noise occurring in the long-term map registration. 

We consider two variants: noisy training and noise-free training. Adding noise to odometry during training additionally affects the reward functions used for policy learning. For \texttt{\small curiosity}, the features used for forward-dynamics prediction are noisy, since they include the noisy occupancy estimates. For \texttt{\small novelty}, the state visitation counts will be noisier since the agent is unsure of its position in the global coordinates. For \texttt{\small coverage}, this directly affects the measure of coverage used in the reward function since coverage measurements rely on accurate odometry. For \texttt{\small reconstruction}, this affects the pose inputs $p^{t}$ associated with each observation obtained during exploration, which may lead to worse predictions. 

Following past work~\cite{chen2019learning}, we use a truncated Gaussian noise model whose standard deviation is proportional to the action magnitudes (see Tab.~1). Specifically, at each time step, we sample a random perturbation from a truncated Gaussian with mean 0, standard deviation $\eta \Delta_{a}$, and a truncation width of $\eta \Delta_{a}$ on either sides of the mean. Here, $\eta$ is the noise level and $\Delta_{a}$ is the action magnitude. The agent estimates its current position by summing up the odometer readings over time. Therefore, the noise gets accumulated over time leading to increasingly large deviations in the agent's pose estimates.

Fig.~\ref{fig:noisy_odometer} compares the four paradigms along with baselines. When trained under noise-free conditions, the learned methods generalize relatively well to different testing noise levels, despite not being exposed to noisy inputs during training. Interestingly, their relative trends are consistent across the different noise levels during testing. However, the performance of \texttt{\small frontier-exploration} deteriorates rapidly as the noise level increases. For noisy training conditions, we set the noise level as $\eta=0.15$. While \texttt{\small coverage} remains robust to noisy training conditions, the exploration performance deteriorates for most learning-based methods when compared to the noise-free training conditions. This is expected since the reward function becomes less reliable under noise-free training conditions. Nevertheless, the learning-based approaches outperform \texttt{\small frontier-exploration} at higher noise levels.

\subsection{Scale factors that influence performance}
\label{sec:analysis}

We now analyze how scaling factors affect exploration quality. 

\paragraph{How does dataset size affect learning?} 
First we analyze how the exploration performance varies with the training dataset size, i.e., the number of unique training environments in MP3D. We select \texttt{\small novelty} and \texttt{\small smooth-coverage} as they are the top performers on MP3D. We additionally select \texttt{\small area-coverage} to compare its behavior with \texttt{\small smooth-coverage}. We train agents on 3 random subsets of $\{10, 20, 40\}$ MP3D environments and measure their exploration performance at $T_{exp}=1000$ on the full MP3D test set. 

\begin{figure}
    \centering
    \includegraphics[width=0.9\textwidth]{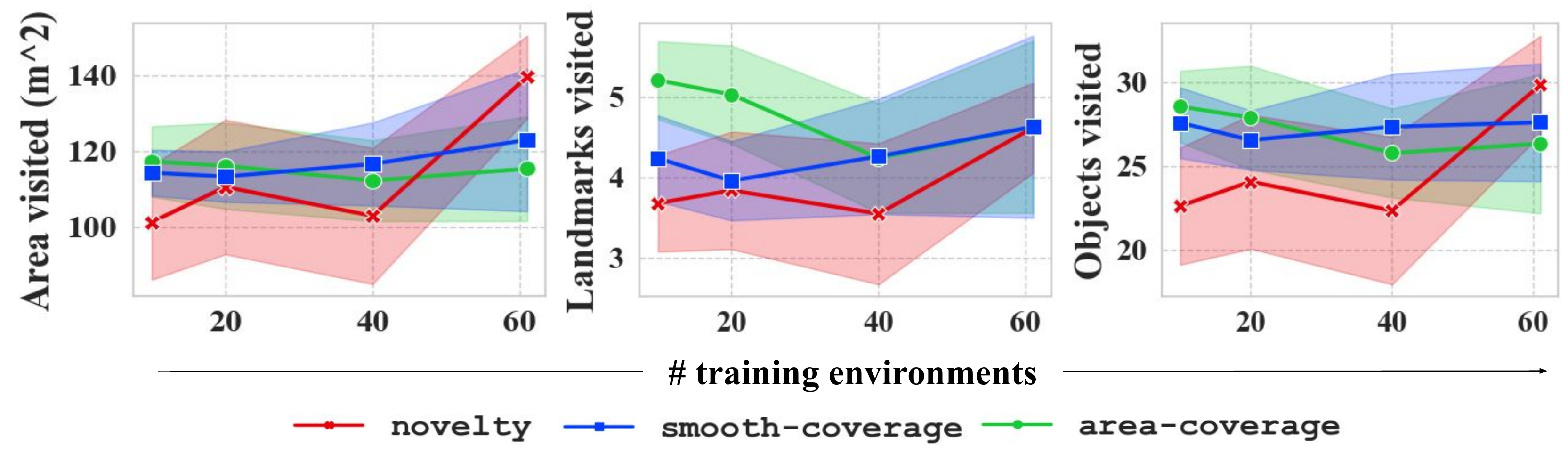}
    \caption{{\footnotesize The impact of varying the number of training environments on exploration performance.}}
    \label{fig:varying_training_envs}
\end{figure}

Fig.~\ref{fig:varying_training_envs} shows the results. Interestingly, all agents achieve very good performance with only 10 training environments. We observe the benefit of well-shaped rewards provided by \texttt{\small novelty} and \texttt{\small smooth-coverage}, which decay exponentially based on the visitation/observation frequency. While their performances improve with the number of environments, it saturates quickly for \texttt{\small area-coverage}. There is gradual but steady improvement in the performance from 10 environments to the full training set, suggesting that these approaches might continue to benefit from significantly larger training environment sets. 

\begin{figure}
    \centering
    \includegraphics[width=0.9\textwidth]{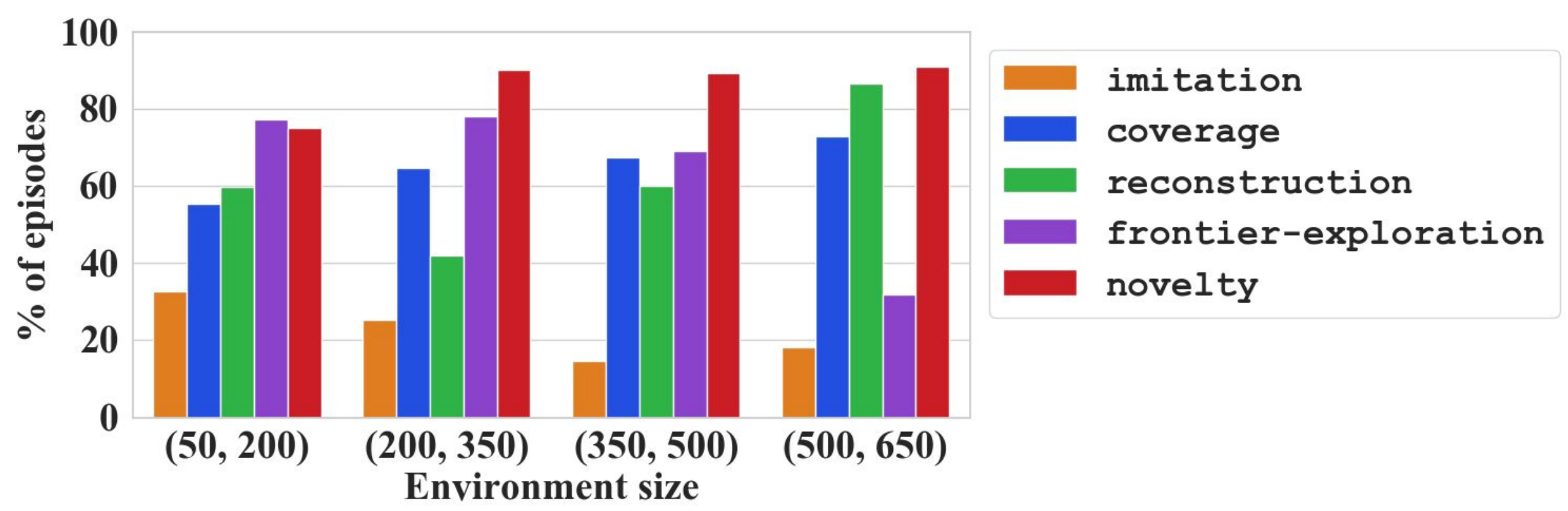}
    \caption{{\footnotesize The impact of varying the testing environment size on exploration performance.}}
    \label{fig:varying_testing_envs}
\end{figure}

\paragraph{How does environment size affect exploration?} 
Next we select the top five methods on MP3D and measure their performance as a function of testing environment size (see Fig.~\ref{fig:varying_testing_envs}). First, we group test episodes based on area visited by the best oracle. Within each group, we report the $\%$ of episodes in which each method ranks in the top 3 out of 5. The larger this value, the better the method is on those environments. The \texttt{\small novelty} and \texttt{\small coverage} approaches are robust and perform well on most environment sizes. The \texttt{\small reconstruction} approach also competes well, especially in larger environments. However, \texttt{\small frontier-exploration} struggles in large MP3D environments as they tend to contain mesh defects where the agent gets stuck. Please see Appendix~\ref{appsec:factors_influencing} for qualitative examples. 

\subsection{Ablation studies}
\label{sec:ablation_studies}
In the foregoing experiments, we have proposed two changes to implementations of prior work: (1) using a smoother variant of the \texttt{\small area-coverage} reward function proposed in~\cite{chen2019learning}, and (2) using the feature representation from a memory architecture for \texttt{\small curiosity} as opposed to image representations as done in~\cite{pathak2017curiosity,burada2018curiosity,chen2019learning}. We now evaluate the effect of these changes on exploration performance.

\paragraph{How do smoother rewards affect exploration?}
As we discussed in Sec.~\ref{sec:coverage}, the formulation of \texttt{\small area-coverage} leads to specific forms of reward sparsity. We addressed this by proposing the \texttt{\small smooth-coverage} method. We now evaluate the impact of making this change across different metrics and datasets. In Tab.~\ref{tab:smooth_coverage}, we compare the two coverage approaches on all six metrics on AVD and MP3D. As we can see, \texttt{\small smooth-coverage} leads to consistent improvements across most metrics on AVD and across the visitation metrics in MP3D. 
\begin{table}[t]
\centering
\scalebox{0.85}{
\begin{tabular}{@{}lcccccc@{}}
     \toprule
                                 &                                                  \multicolumn{6}{c}{Active Vision Dataset}                                                    \\ \cmidrule{2-7}
     \multicolumn{1}{c}{Method}  &        \sm{Area}      &      \sm{Objects}     &     \sm{Landmarks}    &    \sm{View Local.}   &    \sm{Navigation}    &  \sm{Reconstruction}  \\ \midrule
     \texttt{area-coverage}      &\mstd{ 0.85 }{ 0.02 }  & \mstd{ 0.72 }{ 0.00 } & \mstd{ 0.57 }{ 0.01 } & \bstd{ 0.43 }{ 0.02 } & \mstd{ 0.62 }{ 0.01 } & \mstd{ 0.67 }{ 0.05 } \\
     \texttt{smooth-coverage}    &\bstd{ 0.88 }{ 0.02 }  & \bstd{ 0.76 }{ 0.05 } & \bstd{ 0.60 }{ 0.02 } & \mstd{ 0.41 }{ 0.02 } & \bstd{ 0.63 }{ 0.02 } & \bstd{ 0.69 }{ 0.04 } \\ \midrule
                                 &                                                  \multicolumn{6}{c}{Matterport3D}                                                             \\ \cmidrule{2-7}
     \multicolumn{1}{c}{Method}  &        \sm{Area}      &      \sm{Objects}     &     \sm{Landmarks}    &    \sm{View Local.}   &    \sm{Navigation}    &  \sm{Reconstruction}  \\ \midrule
     \texttt{area-coverage}      & \mstd{ 0.54 }{ 0.06 } & \mstd{ 0.52 }{ 0.07 } & \mstd{ 0.34 }{ 0.07 } & \mstd{ 0.20 }{ 0.02 } & \bstd{ 0.66 }{ 0.03 } & \bstd{ 0.44 }{ 0.01 } \\
     \texttt{smooth-coverage}    & \bstd{ 0.58 }{ 0.06 } & \bstd{ 0.56 }{ 0.05 } & \bstd{ 0.35 }{ 0.07 } & \bstd{ 0.21 }{ 0.02 } & \bstd{ 0.67 }{ 0.01 } & \bstd{ 0.44 }{ 0.01 } \\ \bottomrule
\end{tabular}
}
\caption{Ablation study on designing a smoother coverage reward function.}
\label{tab:smooth_coverage}
\end{table}

\begin{figure}[t]
    \centering
    \includegraphics[width=0.9\textwidth]{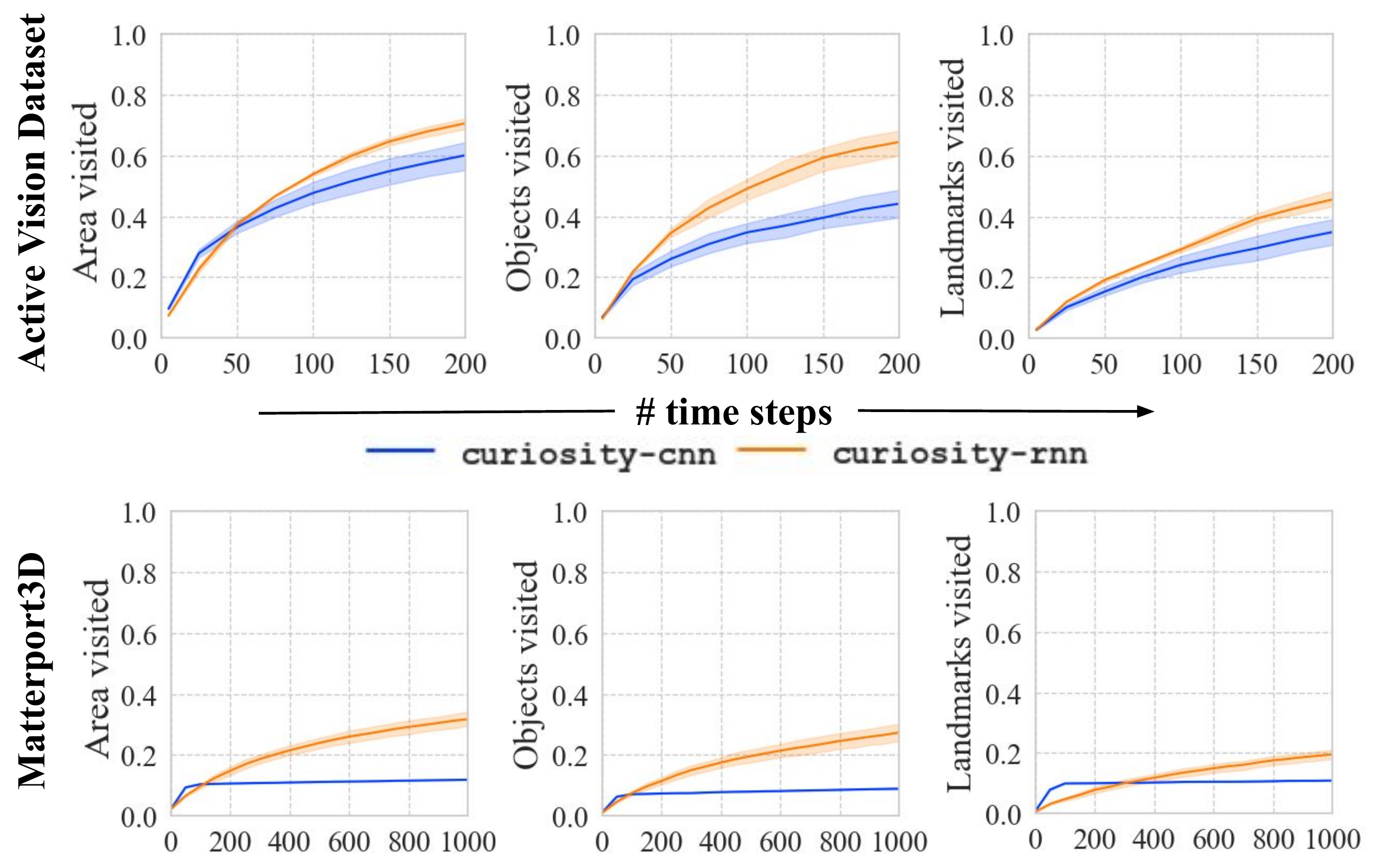}
    \caption{Ablation study on feature representations for curiosity-driven exploration.}
    \label{fig:curiosity_comparison}
\end{figure}

\paragraph{How do memory-based representations influence curiosity-driven exploration?}
As discussed in Sec.~\ref{sec:exploration_results}, curiosity requires good feature representations that account for partial observability in large 3D environments. Using better memory architectures for learning can improve the performance by maintaining longer-term state representations. In Fig.~\ref{fig:curiosity_comparison}, we compare the performance between the spatio-temporal memory (spatial map + RNN) based representation that we use (\texttt{\small curiosity-rnn}) and the version used in~\cite{chen2019learning} where pre-trained ImageNet features were used as the feature representation (\texttt{\small curiosity-cnn}). As we can see, using the RNN feature representation leads to significant improvements on all visitation metrics. In fact, \texttt{\small curiosity-cnn} fails to learn any meaningful behaviors in MP3D. Despite the improvement seen here, the overall performance of \texttt{\small curiosity} is significantly worse than the other paradigms. Curiosity-driven exploration may therefore benefit from having even better memory architectures. In addition, having a large number of parallel environments is critical for training curiosity-based agents~\cite{burada2018curiosity}. This is not practical in our case where we use large-scale 3D environments that have significantly higher computational and data constraints than simpler 2D scenarios. 
\section{Conclusions}
\label{sec:conclusions}

We considered the problem of visual exploration in 3D environments. Prior work presents results on varying experimental conditions, making it hard to analyze what works when. Motivated by this, we proposed a novel benchmark for consistently evaluating exploration algorithms under common experimental conditions: policy architecture, 3D environments, learning algorithm, and diverse evaluation metrics. We then presented a comparative study of four popular exploration paradigms on this standardized benchmark. 

To enable this study, we introduced new metrics and baselines, and we improved upon some existing approaches to scale well to 3D environments. Specifically, we extend the ideas from reconstruction-based exploration $360^\circ$ scene exploration approaches to work on general 3D environments. We also introduced a new coverage reward function that improves upon the existing area-coverage method. Our analysis provides a comprehensive view of the state of the art and each paradigm's strengths and weaknesses. \\

\noindent To recap some of our key findings: 
\begin{itemize}
    \item In the small and cluttered environments from Active Vision Dataset, the \texttt{\small coverage} and \texttt{\small reconstruction} methods are the strongest paradigms as they prioritize selecting views that maximally increase information.
    \item In the large and open environments from Matterport3D, \texttt{\small novelty} and \texttt{\small smooth-coverage} approaches are the strongest paradigms as they have smoother reward functions which are easier to optimize in large environments. 
    \item In the diverse Matterport3D testing environments, different approaches tend to dominate on different skills, highlighting the need for diverse evaluation metrics. 
    \item The performance trends among learned approaches remain consistent in noise-free and noisy test conditions, whereas a purely geometric approach like \texttt{\small frontier-exploration} tends to deteriorate rapidly in the presence of sensor noise. 
    \item Our proposed \texttt{\small smooth-coverage} method improves over a prior coverage approach by using a smoother reward function that eases optimization and leads to consistently better performance on a variety of conditions. 
    \item Our proposed adaptation of \texttt{\small reconstruction} successfully explores 3D environments and competes closely with the best methods on most settings. 
    \item Improved memory architectures may be the key to scaling curiosity-driven exploration to visually-rich 3D environments under extreme partial observability.
    \item An easy-to-implement heuristic \texttt{\small imitation} significantly outperform baselines typically employed, and can serve as a better baseline for future research. 
\end{itemize}

We hope that our study serves as a useful starting point and a reliable benchmark for future research in embodied visual exploration. Code, data, and models are publicly available.

\section*{Acknowledgements}
UT Austin is supported in part by DARPA Lifelong Learning Machines and the GCP Research Credits Program.

\bibliographystyle{spmpsci}
\bibliography{egbib}

\vfill
\pagebreak
\appendix
\appendixpage
\normalsize

\noindent We provide additional information to support the text from the main paper. In particular, the appendix includes additional details regarding the following key topics: \\

\noindent\textbf{Section~\ref{appsec:downstream_task}:} Downstream task transfer \\
\noindent\textbf{Section~\ref{appsec:visitation_criteria}:} Criteria for visiting objects and landmarks \\
\noindent\textbf{Section~\ref{appsec:hyperparameters_exploration}:} Hyperparameters for learning exploration policies \\
\noindent\textbf{Section~\ref{appsec:comparing_coverage_variants}:} Comparative study of different coverage variants \\
\noindent\textbf{Section~\ref{appsec:frontier_exploration}:} Frontier-exploration algorithm \\
\noindent\textbf{Section~\ref{appsec:difficult_pointnav_episodes}:} Generating difficult testing episodes for PointNav \\
\noindent\textbf{Section~\ref{appsec:mining_landmarks}:} Automatically mining landmarks \\
\noindent\textbf{Section~\ref{appsec:factors_influencing}:} The factors influencing exploration performance \\

\section{Downstream task transfer}
\label{appsec:downstream_task}
We now elaborate on the three downstream tasks defined in Sec.~\ref{sec:evaluation_metrics}: \textit{view localization}, \textit{reconstruction}, and \textit{PointNav navigation}.

\subsection{View localization pipeline}
\label{appsec:view_localization_pipeline}

\subsubsection{Problem Setup}
An exploration agent is required to gather information from the environment that will allow it to localize key landmark views in the environment after exploration. Since the exploration agent does not know what views will be presented to it a priori, a general exploration policy that gathers useful information about the environment will perform best on this task.

More formally, the problem of view localization is as follows.  The exploration agent is spawned at a random pose $p_{0}$ in some unknown environment, and is allowed to explore the environment for a time budget $T_{exp}$. Let $V_{exp} = \{x_{t}\}_{t=1}^{T_{exp}}$ be the set of observations the agent received and $\mathcal{P}_{exp} = \{p_{t}\}_{t=1}^{T_{exp}}$ be the corresponding agent poses (relative to pose $p_{0}$). After exploration, a set of $N$ query views $V = \{x_{i}^{q}\}_{i=1}^{N}$ are sampled from query poses $\mathcal{P} = \{p_{i}^{q}\}_{i=1}^{N}$ within the same environment and presented to the agent. The agent is then required to use the information $V_{exp},~\mathcal{P}_{exp}$ gathered during exploration to predict $\{p_{i}^{q}\}_{i=1}^{N}$. In practice, the agent is only required to predict the translation components of the pose, i.e., $p^{ref} = (\Delta x, \Delta y)$ where $\Delta x, \Delta y$ represent the translation along the $X$ and $Y$ axes from a top-down view of the environment. An agent that can successfully predict this has a good understanding of the layout of the environment as it can point to where a large set of views in the environment are sampled from. We next review the architecture for view localization. For the sake of simplicity, we consider the case of $N=1$ with $x^{q}, p^{q}$ denoting the query view and pose respectively.

\subsubsection{View localization architecture}
\begin{figure}[ht]
    \centering
    \includegraphics[trim={0 0cm 0cm 0},clip,width=\textwidth]{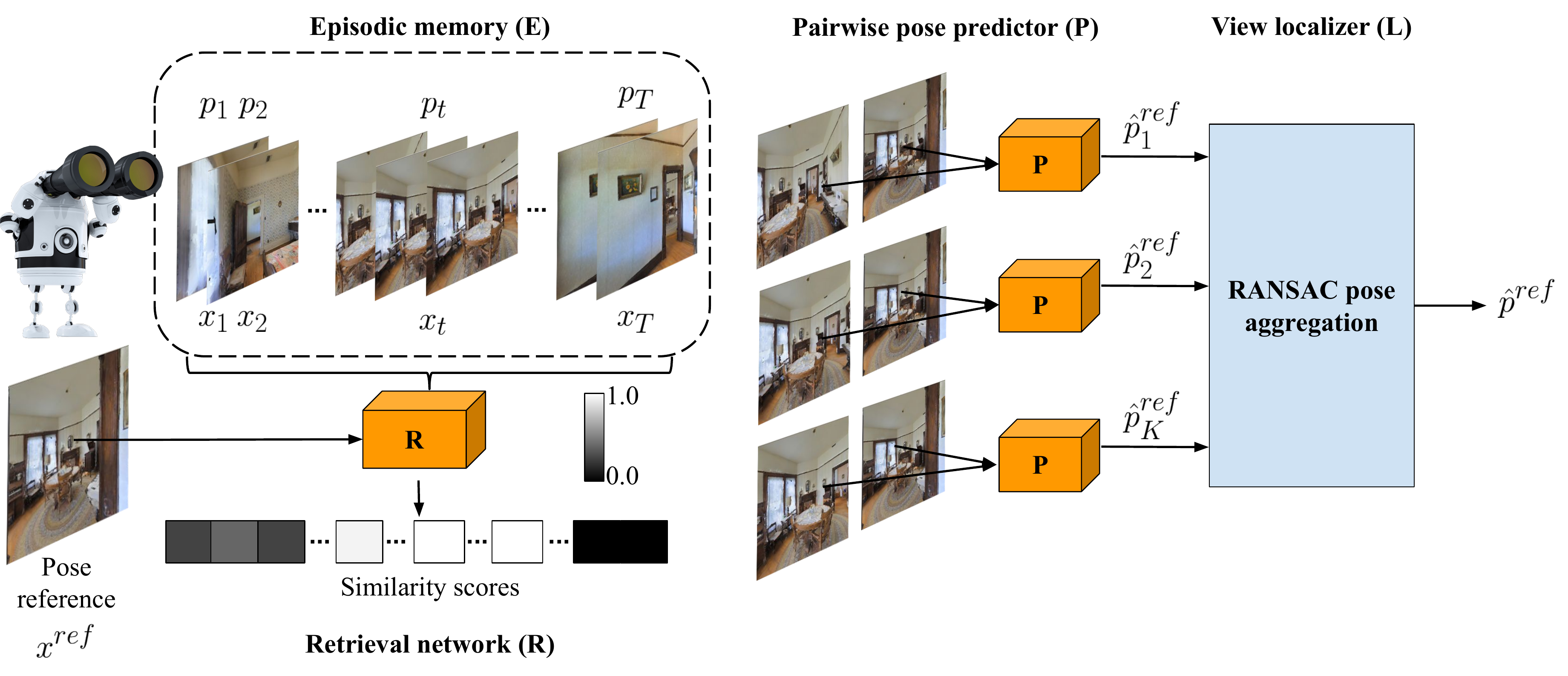}
    \caption{\textbf{View localization architecture:} consists of four main components. \textbf{(1) Episodic memory (E)} (left top) stores the sequences of the agent's past egocentric observations along with their poses (relative to the agent's starting viewpoint). \textbf{(2) Retrieval network (R)} (left bottom) compares a reference image $x^{ref}$ with the episodic memory and retrieves the top $K$ similar images $\{x_{j}\}_{j=1}^{K}$. \textbf{(3) Pairwise pose predictor (P)} (center) estimates the real-world pose $\hat{p}^{ref}_{j}$ of $x^{ref}$ using each retrieval $x_{j}, p_{j}$ and $x^{ref}$. \textbf{(4) View localizer (L)} (right) combines the individual pose predictions $\{\hat{p}^{ref}_{j}\}_{j=1}^{K}$ by filtering the noisy estimates using RANSAC to localize $x^{ref}$.} 
    \label{fig:view_localization_architecture}
\end{figure}

The architecture of our view localization model consists of four components (see Fig.~\ref{fig:view_localization_architecture}). 

\paragraph{Episodic Memory (E)} In order to store information over the course of a trajectory, we use an episodic memory E that stores the history of past observations and corresponding poses $\{(x_{t}, p_{t})\}_{t=1}^{T}$. For efficient storage, we only store image features vectors obtained from the pairwise pose predictor (P) and retrieval network (R) (as described in subsequent sections) in the memory.  

\begin{figure}[ht]
    \centering
    \includegraphics[width=0.25\textwidth,trim={0 3.5cm 10cm 0},clip]{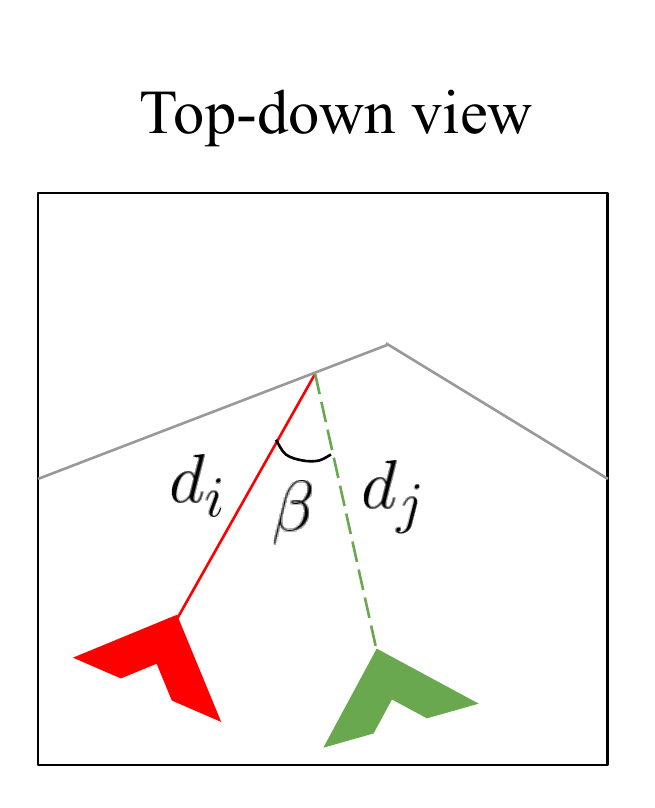}
    \caption{\textbf{Pairwise pose data sampling:} First, a random viewpoint $x_{i}$ (red) is selected from the environment. A ray is cast along its viewing direction to reach the obstacle (gray) at the point of focus. A new ray (green dotted) is cast out from the point of focus and another viewpoint $x_{j}$ (green) is selected along this ray. Since $x_{i}$ and $x_{j}$ share similar visual content, it should be possible to estimate the pose between these viewpoints. $d_{i}, d_{j}$ are the distances from the viewpoints to the point of focus. $\beta$ is the baseline angle between the two viewpoints.}
    \label{fig:pose_sampling}
\end{figure}

\begin{figure}[ht]
    \centering
    \includegraphics[width=0.70\textwidth,trim={0 0 0 0},clip]{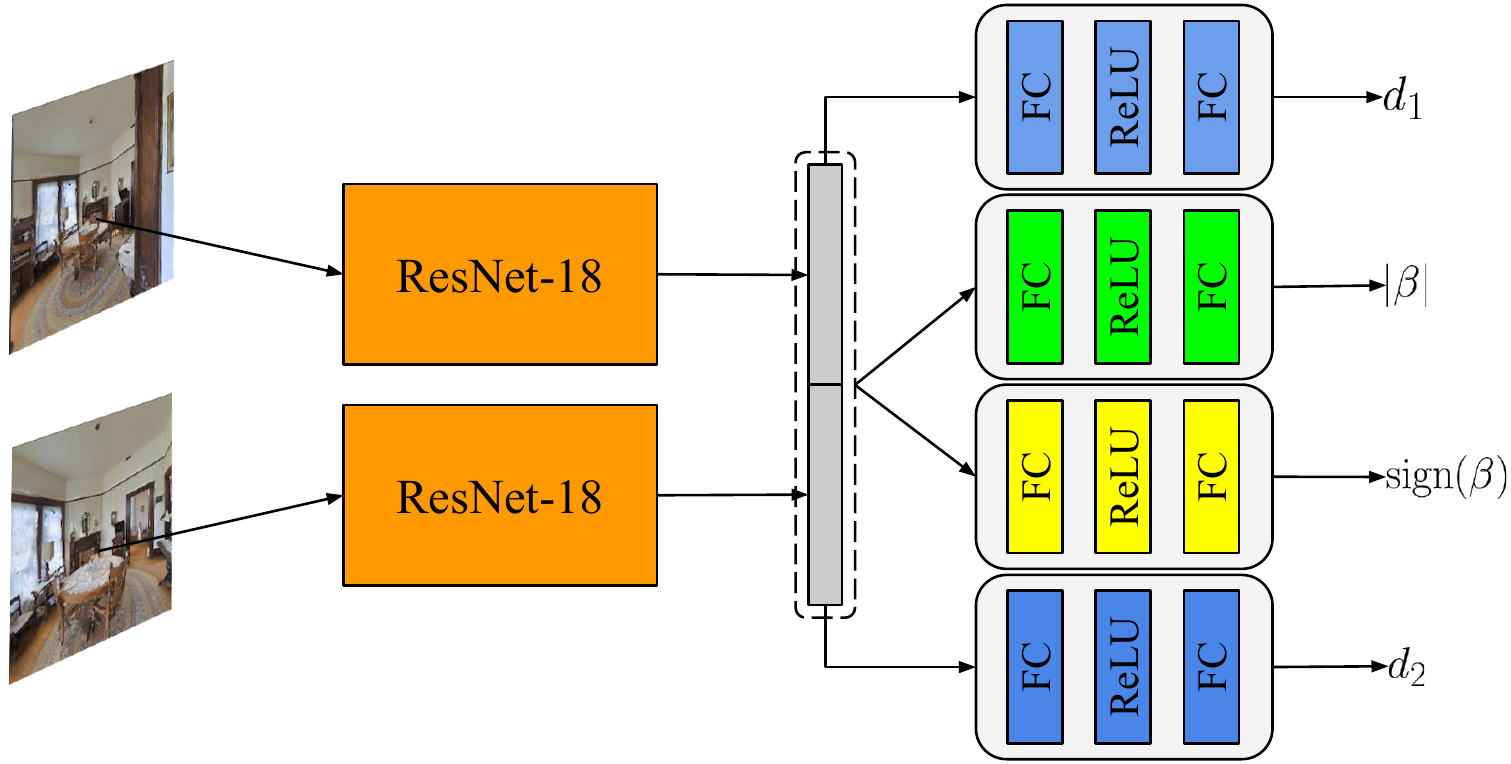}
    \caption{\textbf{Pairwise pose predictor:} A ResNet-18 feature extractor~\cite{he2016deep} extracts features from both images. The concatenated features are then used by three separate networks to predict (1) the distances $d_{1}, d_{2}$ of the points of focus of each image, (2) the magnitude $|\beta|$, and (3) $\textrm{sign}(\beta)$ of the baseline $\beta$ (notations in Fig.~\ref{fig:pose_sampling}). Parameters are shared between the ResNets (orange), and distance prediction MLPs (blue).}
    \label{fig:pairwise_pose_predictor}
\end{figure}

\paragraph{Pairwise pose predictor (P)} We train a pairwise pose predictor that takes in pairs of images $x_{i}, x_{j}$ that are visually similar (see next section), and predicts $\Delta p^{j}_{i} = \textrm{P}(x_{i}, x_{j})$, where $\Delta p^{j}_{i}$ is the relative pose of $x_{j}$ in the egocentric coordinates of $x_{i}$. The architecture is shown in Fig.~\ref{fig:pairwise_pose_predictor}. We follow a different parameterization of the pose prediction when compared to~\cite{zamir2016generic}. Instead of directly regressing $\Delta p^{j}_{i}$, we first predict the distances $d_{i}, d_{j}$ to the points of focus (central pixel) for each image, and the baseline angle $\beta$ between the two viewpoints (see Fig.~\ref{fig:pose_sampling}). The relative pose is then computed as follows:  

\[
    \Delta p_{i}^{j} = (d_{i} - d_{j}\textrm{cos}(\beta),~- d_{j}\textrm{sin}(\beta),~\beta)
\]

This pose parameterization was more effective than directly regressing $\Delta p_{i}^{j}$, especially when the data diversity was limited (eg. in AVD). To sample data for training the pose estimator, we opt for the sampling strategy from~\cite{zamir2016generic} (see Fig.~\ref{fig:pose_sampling}). The prediction of $d_{i}, d_{j}$ is cast as independent regression problems with the MSE loss $L_{d}$. The prediction of $\beta$ is split into two problems: predicting the baseline magnitude, and predicting the baseline sign. Baseline magnitude prediction is treated as a regression problem for AVD and as a 15-class classification problem for MP3D with corresponding MSE or cross entropy losses ($L_{\textrm{mag}})$. Baseline sign prediction is treated as a binary classification problem with a binary cross entropy loss $L_{\textrm{sign}}$. The overall loss function is:

\begin{equation}
    L = L_{d} + L_{\textrm{mag}} + L_{\textrm{sign}}
\end{equation}

\paragraph {Retrieval network (R)} We train a retrieval network R that, given a query image $x^{q}$, can retrieve matching observations from E. Similar to~\cite{savinov2018semi}, we use a siamese architecture consisting of a ResNet-18 feature extractor followed by a 2-layer MLP to predict the similarity score $\textrm{R}(x_{i}, x^{q})$ between images $x_{i}$ and $x^{q}$. Since our goal is to retrieve observations that can be used by the pairwise pose predictor (P), the positive pairs are the same pairs used for training the pose predictor. Negative pairs are obtained by choosing random images that are far away from the current location. We use the binary cross entropy loss function to train the retrieval network.

\paragraph{View localizer (L)} So far, we have a retrieval network R that retrieves observations that are similar to a query view $x^{q}$, and a pairwise pose predictor P that predicts the relative pose between $x^{q}$ and each retrieved image. The goal of the view localizer (L) is to combine predictions made by P on individual retrievals given by R to obtain a robust estimate the final pose $p^{q}$. The overall pipeline works as follows (see Fig.~\ref{fig:view_localization_architecture}).

Similarity scores $\{\textrm{R}(x_{t}, x^{q})\}_{t=1}^{T_{exp}}$ are computed between each $x_{t}$ in the episodic memory E and $x^{q}$. The sequence of scores are temporally smoothed using a median filter to remove noisy predictions. After filtering out dissimilar images in the episodic memory, $\mathcal{V}_{sim} = \{x_{t} | \textrm{R}(x_{t}, x^{q}) < \eta_{noise}\}$, we sample the top $K$ observations $\{x_{j}\}_{j=1}^{K}$ from $\mathcal{V}_{sim}$ with highest similarity scores. For each retrieved observation $x_{j}$, we compute the relative pose $\Delta p^{q}_{j} = \textrm{P}(x_{j}, x^{q})$, i.e., the predicted pose of $x^q$ in the egocentric coordinates of $x_{j}$. We rotate and translate $\Delta p_{j}^{q}$ using $p_{j}$, the real-world pose of $x_{j}$, to get $\hat{p}_{j}^{q}$, the real-world pose of $x^{q}$ estimated from $x_{j}$: 

\begin{equation}
    \hat{p}_{j}^{q} = \bm{R}_{j} \Delta p_{j}^{q} + \bm{t}_{j}
\end{equation}

where $p_{j} = \{\bm{R}_{j}, \bm{t}_{j}\}$ are the rotation and translation components of $p_{j}$. Given the set of individual predictions $\bm{\hat{p}} = \{\hat{p}^{q}_{j}\}_{j=1}^{K}$, we use RANSAC~\cite{fischler1981random} to aggregate these predictions to arrive at a consistent estimate of $\hat{p}^{ref}$.

\subsubsection{Implementation details} 
For AVD, we restrict the baseline angle to lie in the range $[0, 90]^{\circ}$ and depth values to lie in the range $[1.5, 3]\si{m}$. We sample $\sim 1M$ training, $240K$ validation and $400K$ testing pairs. While the number of samples are high, the diversity is quite limited since there are only 20 environments in total. For MP3D, we restrict the baseline angle to lie in the range $[0, 90]^{\circ}$ and depth values to lie in the range $[1, 4]\si{m}$. We sample $\sim 0.5M$ training, $88K$ validation and $144K$ testing pairs. Both the pairwise pose predictor and retrieval network are trained (independently) using Adam optimizer with a learning rate of $0.0001$, weight decay of $0.00001$, batch size of $128$. The ResNet-18 feature extractor is pretrained on ImageNet. The models are trained for 200 epochs and early stopping is performed using the loss on validation data. In case of AVD, the baseline magnitude predictor is a regression model that is trained using MSE loss. In MP3D, the baseline magnitude predictor is a 15-class classification model where each class represents a uniformly sampled bin in the range $[0, 90]^{\circ}$. The choice of classification vs. regression and the number of classes for predicting $|\beta|$ is made based on the validation performance. In both datasets, we use a median filter of size 3 with $\eta_{noise} = 0.95$ for the view localizer L. We sample the reference views from the set of landmark-views that we used in Sec.~\ref{sec:evaluation_metrics}. Since these views are distinct and do not repeat in the environment, they are less ambiguous to localize.

\vfill
\pagebreak
\subsection{Reconstruction pipeline}
\label{appsec:reconstruction_pipeline}

\begin{figure*}[h!]
    \centering
    \includegraphics[width=0.9\textwidth,trim={0 1.0cm 0 0},clip]{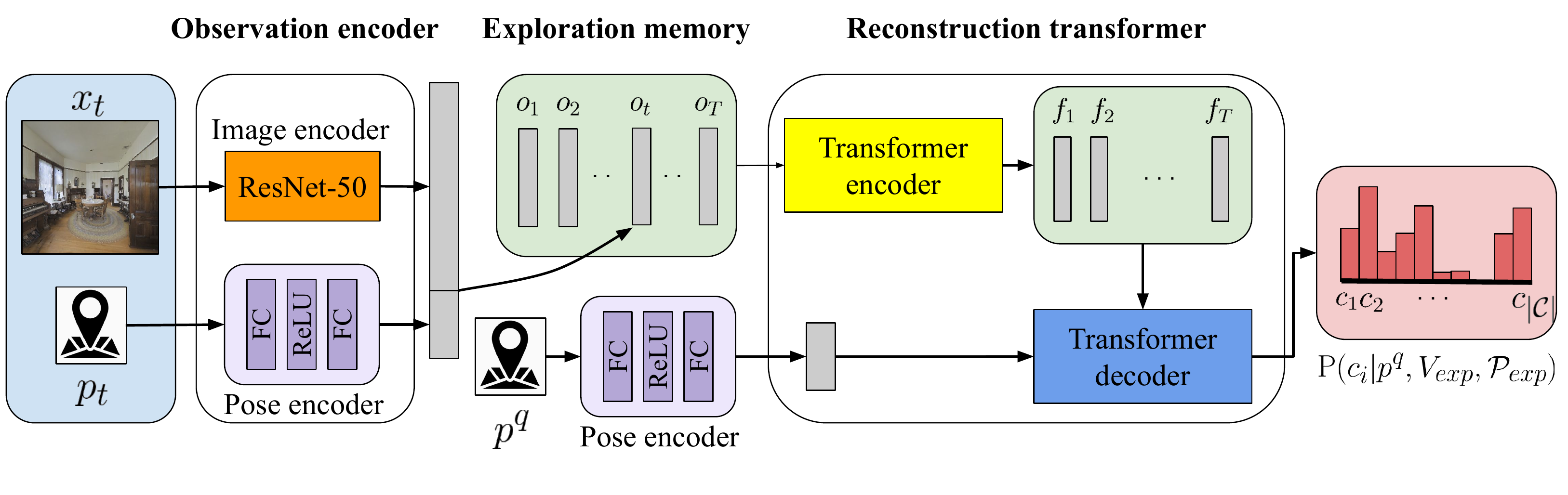}
    \caption{\textbf{Reconstruction architecture:} consists of three key components: \textbf{(1) Observation encoder:} It encodes the input observation $(x_{t}, p_{t})$ obtained during exploration into a high-dimensional feature representation $o_{t}$. The image $x_{t}$ and pose $p_{t}$ are independently encoded using an ImageNet pretrained ResNet-50 and a 2-layer MLP, respectively. \textbf{(2) Exploration memory:} It keeps track of all the encoded features $\{o_{t}\}_{t=1}^{T}$ obtained during exploration, \textbf{(3) Reconstruction transformer:} It contains a transformer encoder and decoder. The transformer encoder uses self-attention between the encoded features to refine the representation and obtain improved features $\mathcal{F} = \{f_{t}\}_{t=1}^{T}$. The transformer decoder uses pose encoding of $p^{q}$ to attend to the right parts of the encoded features $\mathcal{F}$ and predicts a probability distribution over the set of concepts present at a pose $p^{q}$. }
    \label{fig:reconstruction_architecture}
\end{figure*}

\subsubsection{Problem setup: reconstruction}
An exploration agent is required to gather information from the environment that will allow it to reconstruct views from arbitrarily sampled poses in the environment after exploration. Since the exploration agent does not know what poses will be presented to it a priori, a general exploration policy that gathers useful information about the environment will perform best on this task. This can be viewed as the inverse of the view localization problem where views are presented after exploration and their poses must be predicted. \\

Following the task setup from Sec.~\ref{sec:reconstruction} in the main paper, the exploration agent is spawned at a random pose $p_{0}$ in some unknown environment and obtains the observations $V_{exp} = \{x_{t}\}_{t=1}^{T_{exp}}$ views and $\mathcal{P}_{exp} = \{p_{t}\}_{t=1}^{T_{exp}}$ poses during exploration. After exploration, $N$ query poses $\mathcal{P} = \{p_{i}^{q}\}_{i=1}^{N}$ are sampled from the same environment and the agent is required to reconstruct the corresponding views $V = \{x_{i}^{q}\}_{i=1}^{N}$. 

This reconstruction is performed in a concept space $\mathcal{C}$ which is automatically discovered from the training environments. We sample views uniformly from the training environments and cluster their ResNet-50 features using K-means. The concepts $c \in \mathcal{C}$ are, therefore, the cluster centroids obtained after clustering image features. Each query location $p_{i}^{q}$ has a set of ``reconstructing" concepts  $C_{i} = \{c_{i}\}_{i=1}^{J} \in \mathcal{C}$. These are determined by extracting the ResNet-50 features from $x_{i}^{q}$ and obtaining the $J$ nearest cluster centroids. We use a transformer~\cite{vaswani2017attention} based architecture for predicting the concepts as described in Fig.~\ref{fig:reconstruction_architecture}. While this is similar to the policy architecture used in~\cite{fang2019scene}, we use the model to predict concepts present at a given location in the environment instead of learning a motion policy.

\subsubsection{Loss function}
Reconstruction in the concept space is treated as a multilabel classification problem. For a particular query view $x^{q}$ at a query pose $p^{q}$, the reconstructing concepts are obtained by retrieving the top $J$ nearest neighbouring clusters in the image feature space. These $J$ clusters are treated as positive labels for $x^{q}$ and the rest are treated as negative labels. The ground-truth probability distribution $C$ assigned to $x^{q}$ consists of $0$s for the negative labels and $1/J$ for the positive labels. Let $\hat{C} = \textrm{P}(.|p^{q}, V_{exp}, \mathcal{P}_{exp})$ be the posterior probabilites for each concept inferred by the model (see Fig.~\ref{fig:reconstruction_architecture}). Then, the loss $L_{rec}$ is

\[
L_{rec}(p^q) = \infdiv{C}{\hat{C}}
\]

where $D$ is the KL-divergence between the two distributions.

\subsubsection{Reward function}
The \texttt{\small reconstruction} method relies on rewards from a \textit{trained} reconstruction model to learn an exploration policy. Note that the reconstruction model is not updated during policy learning. For each episode, a set of $N$ query poses $\mathcal{P}^{q} = \{p_{i}^{q}\}_{i=1}^{N}$ and their views $V^{q} = \{x_{i}^{q}\}_{i=1}^{N}$ are sampled initially. This information is hidden from the exploration policy and does not affect the exploration directly. At time $t$ during an exploration episode, the agent will have obtained observations $V_{exp}^t = \{x_{\tau}\}_{\tau=1}^{t}$ and $\mathcal{P}_{exp}^t = \{p_{\tau}\}_{\tau=1}^{t}$. The reconstruction model uses $V_{exp}^t, \mathcal{P}_{exp}^t$ to predict posteriors over the concepts for the different queries $p^{q} \in P^{q}$: $\hat{C}_{t}~=~\textrm{P}(.|p^{q}, V_{exp}^t, \mathcal{P}_{exp}^t)$. The reconstruction loss for each prediction is $L_{rec, t}(p^{q}) = \infdiv{C}{\hat{C}_{t}}$ where $C$ is the reconstructing concept set for query $p^q$. The reward is then computed as follows:   

\[
    r_{t} = \frac{1}{N} \sum_{p^q \in \mathcal{P}^{q} }\bigg(L_{rec, t-\Delta_{rec}}(p^q) - L_{rec, t}(p^q)\bigg)
\]

where the reward is provided to the agent after every $\Delta_{rec}$ steps. This reward is the reduction in the reconstruction loss over the past $\Delta_{rec}$ time-steps. The goal of the agent is to constantly reduce the reconstruction loss, and it is rewarded more for larger reductions.

\begin{figure*}[t]
    \centering
    \includegraphics[width=\textwidth]{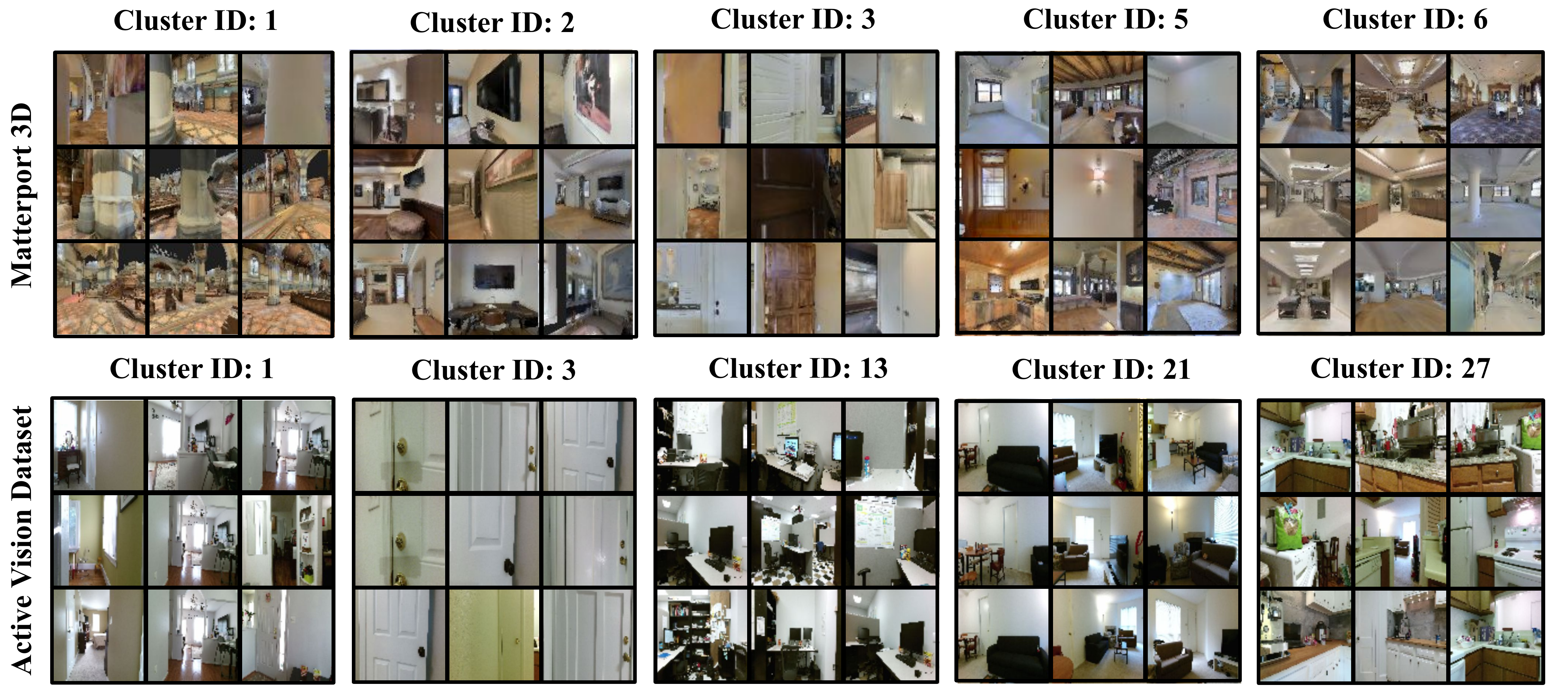}
    \caption{Examples of images in each cluster with the corresponding cluster IDs on Matterport3D (first row) and Active Vision Dataset (second row). The clusters typically corresponding to meaningful concepts such as pillars / arches, doors, windows / lights, geometric layouts in MP3D and windows, doors, computer screens, sofas and kitchen in AVD.}
    \label{fig:reconstruction_clusters}
\end{figure*}

\subsubsection{Implementation details}
We sample $30$ clusters for AVD and $50$ clusters for MP3D based on the Elbow method which selects the $N$, after which, the reduction in within-cluster separation saturates. Additionally, we manually inspect the clusters for different values of $N$ to ensure that they contain meaningful concepts (see Fig.~\ref{fig:reconstruction_clusters}). \\

In practice, we do not directly use the ResNet-50 image features as the output of the image encoder. We compute the similarity scores between the ResNet-50 features for a given $x_{t}$ and all the cluster centroids in $\mathcal{C}$. This gives an $30$ and $50$ dimensional vectors of similarities for AVD and MP3D, respectively which serves as the output of the image encoder. This design choice achieves two things, (1) it reduces computational complexity significantly as the number of clusters is much fewer than the ResNet-50 features ($2048$-D), and (2) it directly incorporates the information from cluster centroids into the reasoning process of the reconstruction, as the reasoning happens in the cluster similarity space rather than image feature space. \\
    
For training the reconstruction model, we sample exploration trajectories using the \texttt{\small oracle} exploration method. First, $N$ query views are sampled for each environment by defining a discrete grid of locations and sampling images from multiple heading angles at each valid location on the grid. For AVD, we use a grid cell distance of $1\si{m}$ while sampling views from $4$ uniformly separated heading angles. For MP3D, we use a grid cell distance of $2\si{m}$ while sampling views from $3$ uniformly separated heading angles. These values were selected to ensure a good spread of views, low redundancy in the views and adequate supervision (larger the grid cell distance, lesser the number of valid points). The model is trained on trajectories of length $T_{exp} = 200$ in AVD and $T_{exp} = 500$ in MP3D. We use $J = 3$ nearest neighbors clusters as positives for both AVD and MP3D. For making the model more robust to the actual trajectory length, we also train on intermediate time-steps of the episode (after every $20$ steps in AVD and $100$ steps in MP3D). The optimization was performed using Adam optimizer with a learning rate of $0.0001$ for AVD and $0.00003$ for MP3D. We use 2 layers in both the transformer encoder and decoder with 2 attention heads each. For training the \texttt{\small reconstruction} exploration agent, we use $\Delta_{rec} = 1$ for AVD and $\Delta_{rec} = 5$ for MP3D.

\vfill
\pagebreak
\subsection{Navigation pipeline}
\label{appsec:navigation_pipeline}

\subsubsection{Problem setup}
An exploration agent is required to gather information from the environment that will allow it to navigate to a given $p^{tgt}$ location after exploration. \\
More formally, the exploration agent is spawned at a random pose $p_{0}$ in some unknown environment, and is allowed to explore the environment for a time budget $T_{exp}$. Let $V^{d}_{exp} = \{x_{t}^{d}\}_{t=1}^{T_{exp}}$ be the set of depth observations the agent received and $\mathcal{P}_{exp} = \{p_{t}\}_{t=1}^{T_{exp}}$ be the corresponding agent poses (relative to pose $p_{0}$). The depth observations along with the corresponding poses are used to build a 2D top-down occupancy map of the environment $\mathcal{M} \in \mathbb{R}^{h\times w}$ that indicates whether an $(x, y)$ location in the map is free, occupied, or unknown. After exploration, the agent is respawned at $p_{0}$ and is provided a target coordinate $p^{tgt}$ that it must navigate to within a budget of time $T_{nav}$, using the occupancy information $\mathcal{M}$ gathered during exploration. After reaching the target, it is required to execute a STOP action indicating that it has successfully reached the target. Following past work on navigation~\cite{anderson2018evaluation,habitat19iccv}, the episode is considered to be a success only of the agent executed the stop action within a threshold geodesic distance $\eta_{\textrm{success}}$ from the target.

\subsubsection{Navigation policy}
We perform navigation using an A* planner that generates a path to the target at each time-step (See Algo.~\ref{algo:navigation_policy}). The input to the policy consists of the egocentric occupancy map $\mathcal{M}$ generated at the end of exploration and a target location $p_{tgt}$ on that map. The map $\mathcal{M}$ consists of free, occupied and unexplored regions. $\textrm{ProcessMap}(\mathcal{M})$ converts this into a binary map by treating all free and unexplored regions as free space, and the occupied regions as obstacles. It also applies the morphology close operator to fill any holes in the binary map. 

Next, the AStarPlanner uses the processed map $\bar{\mathcal{M}}$ to generate the shortest path from the current position to the target. If the policy has reached the target, then it returns STOP. Otherwise, if the path is successfully generated, the policy samples the next location on the path to navigate to ($p^{\textrm{next}}$) and selects an action to navigate to that target. $\textrm{get\_action}()$ is a simple rule-based action selector that moves forward if the agent is already facing the target, otherwise rotates left / right to face $p^{\textrm{next}}$. However, if the path does not exist, the policy samples a random action. This condition is typically reached if ProcessMap blocks narrow paths to the target or assigns the agent's position as an obstacle while closing holes. 

\begin{algorithm}[]
\SetAlgoLined
\KwData{Map $\mathcal{M}$, target $p^{tgt}$}
$\bar{\mathcal{M}} = \textrm{ProcessMap}(\mathcal{M})$\;

$\textrm{Path}_{tgt} = \textrm{AStarPlanner}(\bar{\mathcal{M}}, p^{tgt})$\;
\uIf{\upshape Reached $p^{tgt}$}{
    $a = \textrm{STOP}$;
}
\uElseIf{\upshape $\textrm{Path}_{tgt}$ $\textrm{is not None}$}{
    $p^{\textrm{next}} = \textrm{Path}_{tgt}[\Delta_{\textrm{next}}]$; \\ 
    $a = \textrm{get\_action}(p^{\textrm{next}})$;
}
\Else{
    $a = \textrm{random\_action}()$;
}
\caption{Navigation policy}
\label{algo:navigation_policy}
\end{algorithm}

\subsubsection{Implementation details}
We use a publicly available A* implementation: {\small \url{https://github.com/hjweide/a-star}}. We vary $T_{exp}$ for benchmarking and set $T_{nav} = 200, 500$ for AVD, MP3D. $\eta_{success} = 0.5\si{m}, 0.25\si{m}$ for AVD, MP3D. The value is larger for AVD since the environment is discrete, and a threshold of $0.5\si{m}$ is satisfied only when the agent is one-step away from the target. 

The map $\mathcal{M}$ is an egocentric crop of the allocentric map generated during exploration. For AVD, we freeze the map after exploration, i.e., do not update the map based on observations received during the navigation phase. Therefore, the agent is required to have successfully discovered a path to the target during exploration (eventhough it does not know the target during exploration). This type of evaluation generally fails for MP3D since the floor-plans are very large and it is generally not possible for an exploration agent to discover the full floor plan within the restricted time-budget. Therefore, we permit online updates to $\mathcal{M}$ during exploration. This means that the role of exploration in MP3D is to not necessarily discover a path to the target, instead, it is used to rule out certain regions of the environment that may cause planning failure, which would reduce the navigation efficiency.

\section{Criteria for visiting objects and landmarks}
\label{appsec:visitation_criteria}
We highlight the exact success criteria for what counts as visiting an object or landmark.

\subsection{Visiting objects}
\paragraph{AVD} The object instances in AVD~\cite{ammirato2016avd} are annotated as follows: If an object is visible in an image, the bounding box and the instance ID are listed. A particular object instance is considered to be visited if it is annotated in the current image, the distance to the object is lesser than $1.5\si{m}$, and the bounding box area is larger than 70 squared pixels (approximately $1\%$ of an $84 \times 84$ image). We keep the bounding box size threshold low since many of the object instances in AVD are very small objects\footnote{Examples of AVD object instances: \url{https://www.cs.unc.edu/~ammirato/active_vision_dataset_website/get_data.html}}, and we primarily rely on visibility and the agent's proximity to the object to determine visitation. 

\paragraph{MP3D} The objects in MP3D~\cite{chang2017matterport} are annotated with $(x, y, z)$ center-coordinates in 3D space along with their extents specified as (width, height, depth). As stated in Sec.~\ref{sec:evaluation_metrics} in the main paper, to determine object visitation, we check if the agent is close to the object, has the object within its field of view, and if the object is not occluded. While it is also possible to arrive at similar visitation criteria by rendering semantic segmentations of the scene at each time step, we refrain from doing that as it typically requires larger memory to load semantic meshes and slows down rendering significantly. See Fig.~\ref{fig:object_visitation} for the exact criteria. The values for this evaluation metric were determined upon  manual inspection on training environments.

\begin{table}[h!]
\begin{minipage}{\linewidth}
    \centering
    \includegraphics[width=0.45\textwidth,trim={0 3cm 3cm 0.5cm},clip]{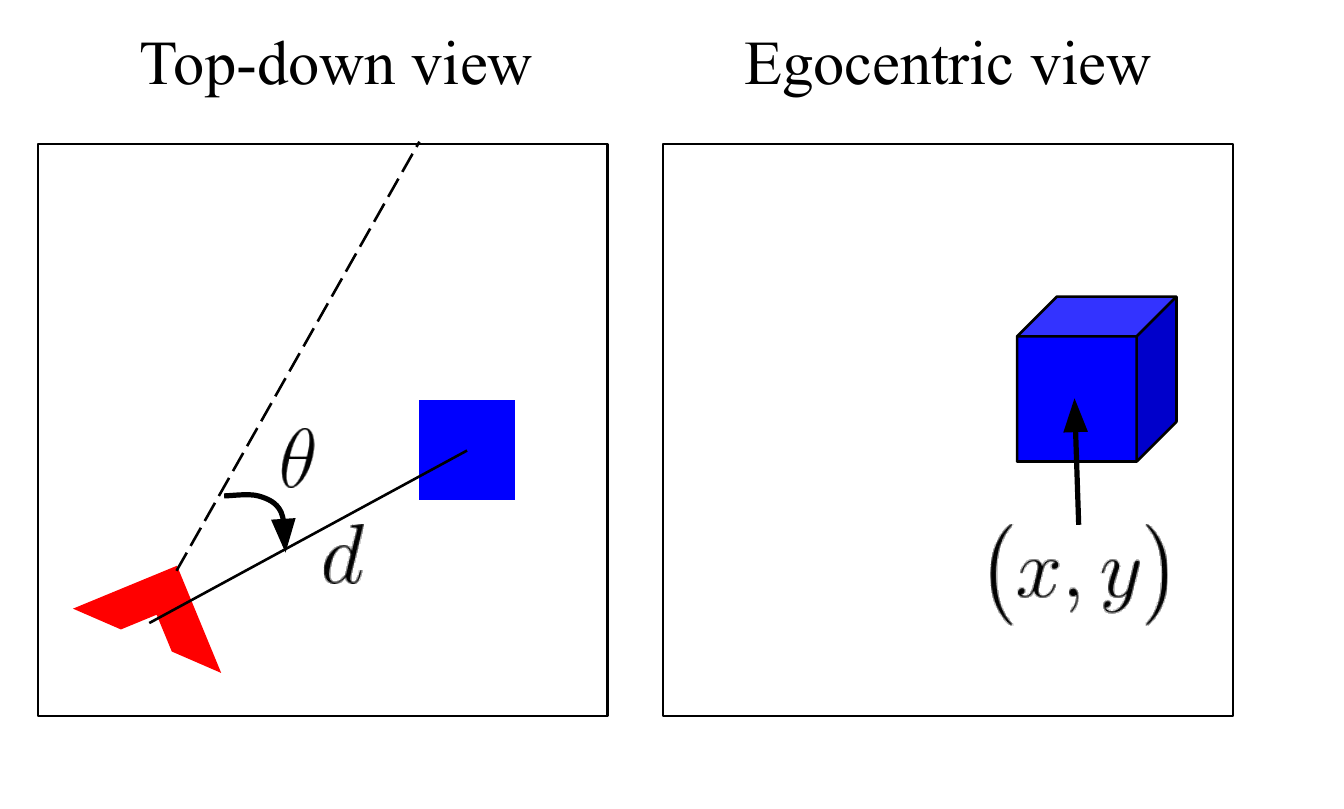}
    \captionof{figure}{\textbf{Object visitation criteria on MP3D:} The left image shows a top-down view of the environment containing the agent (red) and the object (blue). $d$ is the euclidean distance between the agent's centroid and the object's centroid, the dotted line represents the center of the agent's field of view, and $\theta$ represents the angle between the agent's viewing angle and the ray connecting the agent's centroid to the object's centroid. The right image shows the egocentric view of the agent containing the blue object. The indicated $(x, y)$ represents the pixel coordinates of the object centroid. An object is considered visited if (1) $d < 3.0\si{m}$, (2) $\theta \le 60^{\circ}$, (3) $(x, y)$ is within the image extents, and (4) $|\textrm{depth}[x, y] - d\textrm{cos}(\theta)| < 0.3\si{m}$ where $\textrm{depth}$ is the depth image. The final criteria checks for occlusions since the expected distance to the object must be consistent with the depth sensor readings at the object centroid.}
    \label{fig:object_visitation}
\end{minipage}
\begin{minipage}{\linewidth}
    \centering
    \includegraphics[width=0.7\textwidth,trim={0 1.5cm 0 0},clip]{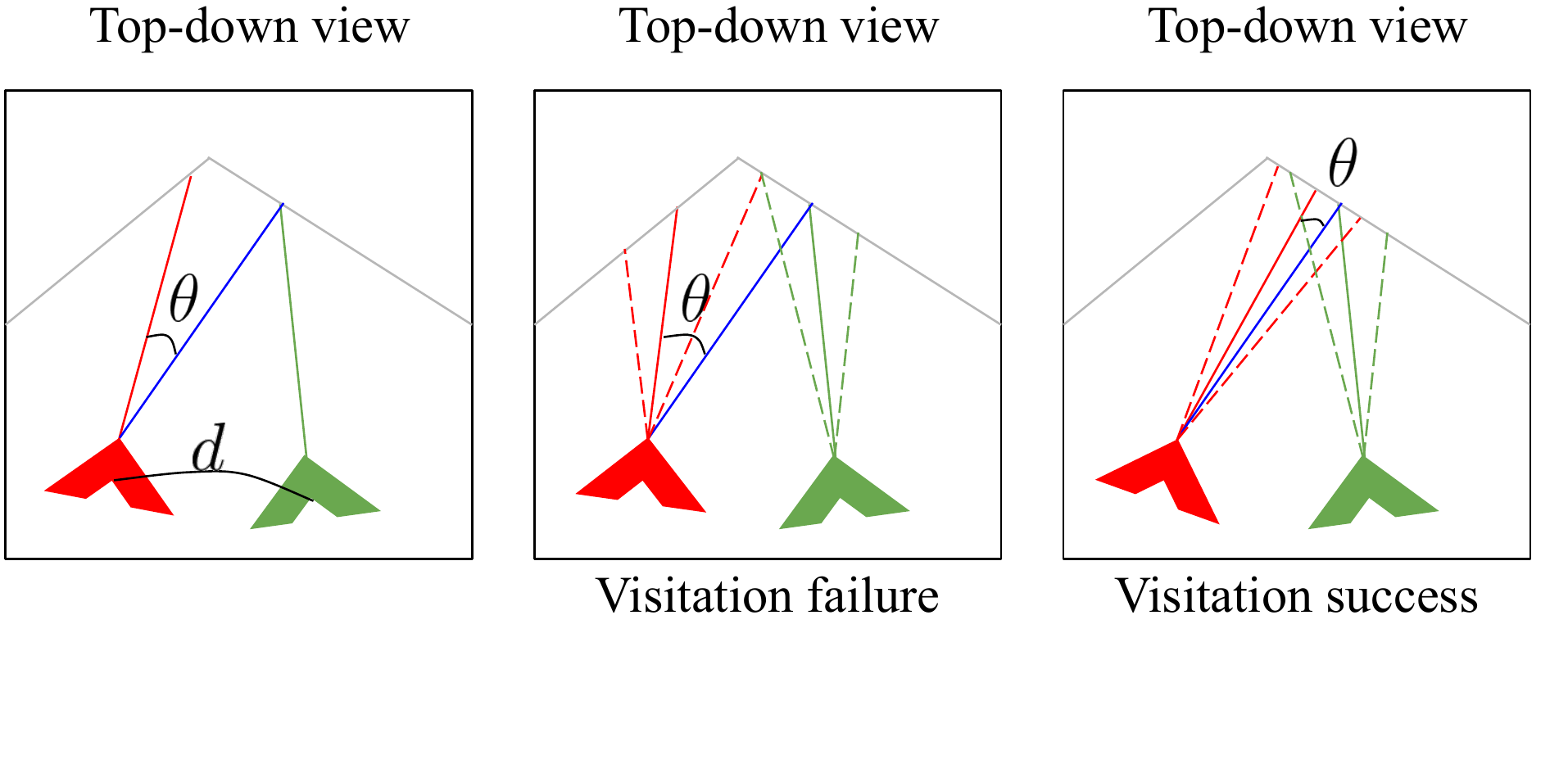}
    \captionof{figure}{\textbf{Landmarks visitation criteria on AVD, MP3D:} The image on the left shows the top-down view of the environment with the agent in red, the landmark-view in green, and rays representing their field of view centers in corresponding colors. The gray lines represent obstacles. $\theta$ represents the discrepancy between the direction the agent is looking at and the point the landmark-view is focused on. $d$ is the geodesic distance between the two viewpoints. To successfully visit the landmark-view, the field of view of the agent must closely overlap with that of the landmark-view. On AVD, this is ensured by satisfying two criteria: (1) $\theta < 30^{\circ}$, and (2) $d < 1\si{m}$. On MP3D, we specify three criteria: (1) $\theta < 20^{\circ}$, (2) $d < 2\si{m}$, and (3) $|d_{1} - d_{2}| < 0.5\si{m}$ where $d_{1}, d_{2}$ are the lengths of the red and blue line-segments respectively. We additionally imposed the third condition to check for occlusions that block agent's view of the landmarks. If the agent is close to the landmark-view, lower $\theta$ leads to success.}
    \label{fig:landmarks_visitation}
\end{minipage}
\end{table}

\subsection{Visiting landmarks}
The criteria for visiting landmarks differs from that of visiting objects as the goal here is to match a particular $(x, y, z, \phi)$ pose in the environment rather than be close to some $(x, y, z)$ location and have it within the agent's field of view. Specifically, the goal is to look at the \textit{same things} that the landmarks are looking at. See Fig.~\ref{fig:landmarks_visitation}.

\section{Hyperparameters for learning exploration policies}
\label{appsec:hyperparameters_exploration}

\begin{table}[h!]
\centering
\begin{tabular}{lcc}
                                                          & AVD                 & MP3D                \\ 
\toprule    
\multicolumn{3}{c}{Optimization}                                                                      \\ \midrule
\multicolumn{1}{l}{Optimizer}                             & Adam                & Adam                \\
\multicolumn{1}{l}{Learning rate}                         & 0.00001 - 0.001     & 0.00001 - 0.001     \\
\multicolumn{1}{l}{$\#$ parallel actors}                  & 32                  & 8                   \\
\multicolumn{1}{l}{PPO mini-batches}                      & 4                   & 2                   \\
\multicolumn{1}{l}{PPO epochs}                            & 4                   & 4                   \\
\multicolumn{1}{l}{Training episode length}               & 200                 & 500                 \\
\multicolumn{1}{l}{GRU history length}                    & 50                  & 100                 \\
\multicolumn{1}{l}{$\#$ training steps (in millions)}     & 12.8                & 8                   \\ \midrule
\multicolumn{3}{c}{Spatial memory}                                                                    \\ \midrule    
\multicolumn{1}{l}{Map bin size $s$}                      & 0.05m               & 0.1m                \\
\multicolumn{1}{l}{$\eta_{l}$}                            & 0.3m                & 0.5m                \\
\multicolumn{1}{l}{$\eta_{h}$}                            & 1.8m                & 2.0m                \\
\multicolumn{1}{l}{$S_{\textrm{coarse}}$}                 & 10.0m               & 20.0m               \\
\multicolumn{1}{l}{$S_{\textrm{fine}}$}                   & 3.0m                & 4.0m                \\ \midrule
\multicolumn{3}{c}{Reward scaling factors for different methods}                                      \\ \midrule
\multicolumn{1}{l}{Method}                                & AVD                 & MP3D                \\ \midrule
\multicolumn{1}{l}{\texttt{\small curiosity}}             & 0.001               & 0.0001              \\
\multicolumn{1}{l}{\texttt{\small novelty}}               & 0.1                 & 0.1                 \\
\multicolumn{1}{l}{\texttt{\small smooth-coverage}}       & 0.3                 & 0.3                 \\
\multicolumn{1}{l}{\texttt{\small reconstruction}}        & 0.1                 & 1.0                 \\
\multicolumn{1}{l}{\texttt{\small area-coverage}}         & 0.01                & 0.001               \\
\multicolumn{1}{l}{\texttt{\small random-views-coverage}} & 1.0                 & 0.3                 \\
\multicolumn{1}{l}{\texttt{\small landmarks-coverage}}    & 1.0                 & 1.0                 \\
\multicolumn{1}{l}{\texttt{\small objects-coverage}}      & 1.0                 & 0.3                 \\ 
\bottomrule
\end{tabular}
\caption{Values for hyperparameters for optimizing exploration policies and the spatial memory common across methods. The learning rate is selected from the specified range based on grid-search. }
\label{tab:exp_hyperparams}
\end{table}
We expand on the implementation details provided in Sec.~\ref{sec:experiments} from the main paper. We use PyTorch~\cite{paszke2017automatic} and a publicly available codebase for PPO~\cite{pytorchrl} for all our experiments. The hyperparameters for training different exploration algorithms are shown in Tab.~\ref{tab:exp_hyperparams}. The optimization and spatial memory hyperparameters are kept fixed across different exploration algorithms. The primary factor that varies across methods is the reward scale. For MP3D, the models are trained on 4 Titan V GPUs and typically take 1-2 days for training. For AVD, the models are trained on 1 Titan V GPU and typically take 1 day to train.

Next, we compare our \texttt{\small curiosity} implementation with the one given in~\cite{burada2018curiosity}. For training our \texttt{\small curiosity} policy, we use the forward-dynamics architecture proposed in~\cite{burada2018curiosity} which consists of four MLP residual blocks.  We use the GRU hidden state from the policy as our feature representation to account for partial observability. As recommended in~\cite{burada2018curiosity}, we do not backpropagate the gradients from the forward dynamics model to the feature representation to have relatively stable features. However, since the policy is updated, the features are not fixed during training (as suggested in~\cite{burada2018curiosity}). Nevertheless, we found that it was more important to use memory-based features that account for partial observability, than to use stable image features that are frozen (see Fig.~\ref{fig:curiosity_comparison} in the main paper). We additionally use PPO, advantage normalization, reward normalization, feature normalization, and observation normalization following the practical considerations suggested in~\cite{burada2018curiosity}. We are limited to using only 8 parallel actors due to computational and memory constraints.

\section{Comparative study of different coverage variants}
\label{appsec:comparing_coverage_variants}
While we use area as our primary quantity of interest for coverage in the main paper, we extend this idea can more generally  for learning to visit other interesting things such as objects (similar to the search task from~\cite{fang2019scene}) and landmarks (see Sec~\ref{sec:evaluation_metrics} from the main paper).  The \texttt{\small coverage} reward consists of the increment in some observed quantity of interest: 
\begin{equation}
    r_{t} \propto I_{t} - I_{t-1},
\end{equation}
where $I_{t}$ is the quantity of interesting things (e.g., object) visited by time $t$. Apart from area coverage (regular and smooth), we also consider objects and landmarks for $I$, where the agent is reward based on the corresponding visitation metric from Sec.~\ref{sec:evaluation_metrics} in the main paper.
\begin{figure}
    \centering
    \includegraphics[width=\textwidth]{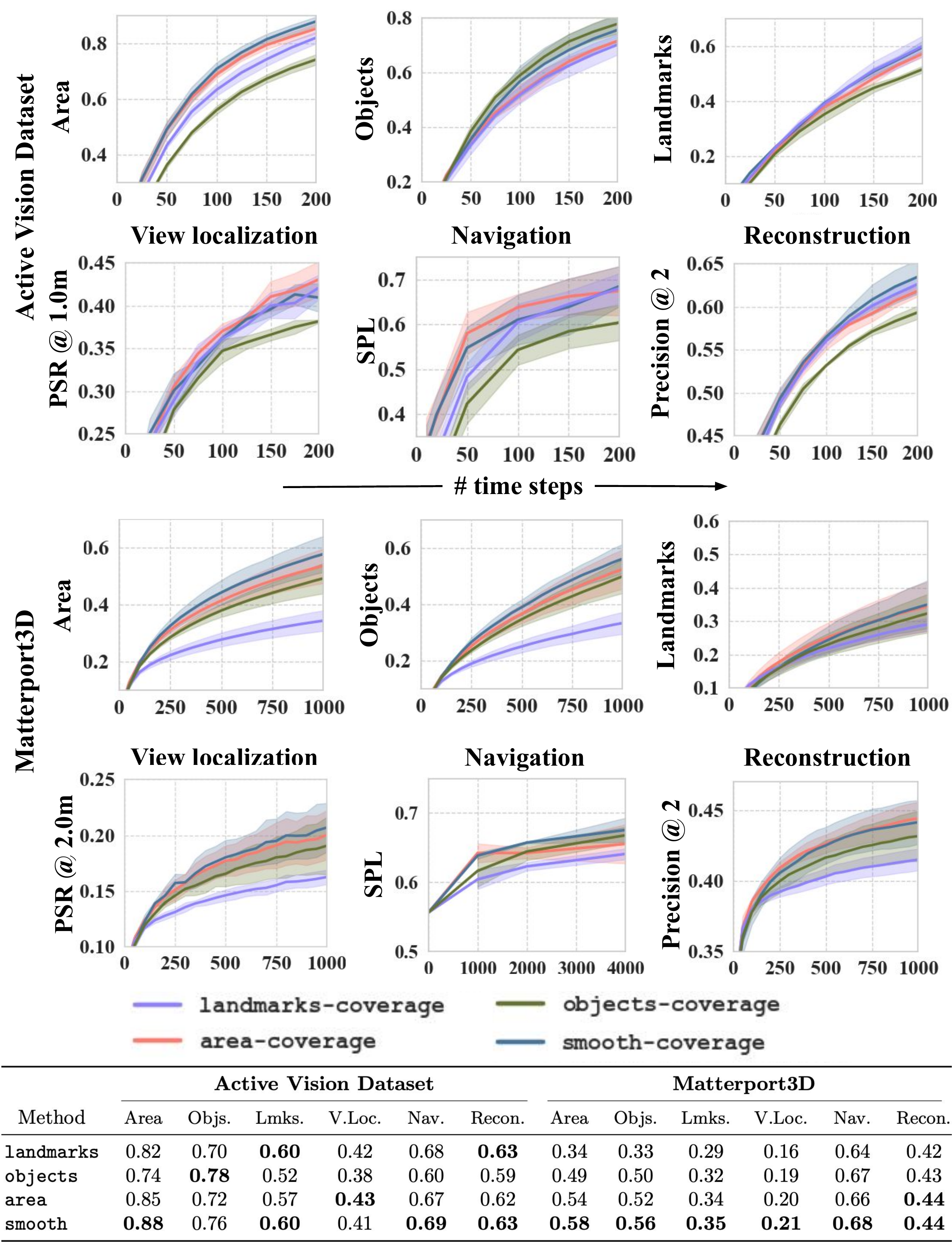}
    \caption{Plots comparing different coverage variants on the three visitation metrics.}
    \label{fig:highres_coverage_visitation_plots}
\end{figure}

For each of the visitation metrics, we have one method that is optimized for doing well on that metric. For example, \texttt{\small area-coverage} optimizes for area visited, \texttt{\small objects-coverage} optimizes for objects visited, etc. As expected, on AVD we generally observe that the method optimized for a particular metric typically does better than most methods on that metric. However, on MP3D, we find that \texttt{\small smooth-coverage} and \texttt{\small area-coverage} dominate on most metrics. This shift in the trend is likely due to optimization difficulties caused by reward sparsity: landmarks and objects occur more sparsely in the large MP3D environments. Objects tend to occur more frequently in the environment than landmarks, and this is reflected in the performance as \texttt{\small objects-coverage} generally performs better. For this reason, we use \texttt{\small smooth-coverage} as the standard coverage method in the main paper.

\vfill
\pagebreak
\section{Frontier-based exploration algorithm}
\label{appsec:frontier_exploration}

We briefly describe the \texttt{\small frontier-exploration} baseline in Sec.~\ref{sec:baselines} in the main paper. Here, we provide a detailed description of the algorithm along with the pseudo-code.  \\

We implement a variant of the frontier-based exploration algorithm from~\cite{yamauchi1997frontier} as shown in Algo.~\ref{algo:frontier_policy}. The core structure of the algorithm is similar to the navigation policy used in Algo.~\ref{algo:navigation_policy}. The key difference here is that the target $p^{tgt}$ is assigned by the algorithm itself. 

\vspace{-0.4cm}
\paragraph{DetectFrontiers()} Given the egocentric occupancy map $\mathcal{M}$ of the environment, frontiers are detected. Frontiers are defined as the edges between free and unexplored regions in the map. In our case, we detect these edges and group them into contours using the contour detection algorithm from OpenCV. ``frontiers" is the list of these contours representing different frontiers in the environment. 

\vspace{-0.4cm}
\paragraph{SampleTarget()} We sort the frontiers based on their lengths since longer contours represent potentially larger areas to uncover. 
We then randomly sample one of the three longest frontiers, and sample a random point within this contour to get $p^{tgt}$.

\vspace{-0.4cm}
\paragraph{UpdateMap()} We update the map based on observations received while navigating to $p^{tgt}$. Once we sample a frontier target $p^{tgt}$, we use a navigation policy (see Algo.~1 in the main paper) to navigate to the target. Since the occupancy maps can be noisy, we add two simple heuristics to make frontier-based exploration more robust. First, we keep track of the number of times planning to $p^{tgt}$ fails. This can happen if the map that is updated during exploration reveals that it is not possible to reach $p^{tgt}$. Second, we keep track of the total time spent on navigating to $p^{tgt}$. Depending on the map updates during exploration, certain targets may be very far away from the agent's current position since the geodesic distance changes based on the revealed obstacles. If planning fails more than $N_{fail}$ times or the time spent reaching $p^{tgt}$ crosses $T_{max}$, then we sampled a new frontier target.

We use $N_{fail} = 2$ for both AVD, MP3D and $T_{max} = 20, 200$ for AVD and MP3D respectively. 
Preliminary analysis showed that the algorithm was relatively robust to different values of $T_{max}$. 

\begin{algorithm}[t]
\SetAlgoLined
\KwData{Map $\mathcal{M}$, maximum time per target $T_{max}$, maximum failures per target $N_{fail}$}
\While{exploration budget not reached}
{
    $\textrm{frontiers} = \textrm{DetectFrontiers}(\mathcal{M})$\;
    $p^{tgt} = \textrm{SampleTarget}(\textrm{frontiers})$\;
    $\textrm{failure\_count} = 0$\;
    $\textrm{time\_spent} = 0$\;
    
    \While{not reached $p^{tgt}$}{
        UpdateMap($\mathcal{M}$)\;
        $\bar{\mathcal{M}} = \textrm{ProcessMap}(\mathcal{M})$\;
        $\textrm{Path}_{tgt} = \textrm{AStarPlanner}(\bar{\mathcal{M}}, p^{tgt})$\;
        \uIf{\upshape Reached $p^{tgt}$}{
            break;
        }
        \uElseIf{\upshape $\textrm{failure\_count} > N_{fail}$ or $\textrm{time\_spent} > T_{max}$}{
            break;
        }
        \uElseIf{\upshape $\textrm{Path}_{tgt}$ $\textrm{is not None}$}{
            $p^{\textrm{next}} = \textrm{Path}_{tgt}[\Delta_{\textrm{next}}]$; \\ 
            $a = \textrm{get\_action}(p^{\textrm{next}})$;
        }
        \Else{
            $a = \textrm{random\_action}()$\;
            $\textrm{failure\_count +=} 1$\;
        }
        $\textrm{time\_spent +=} 1$\;
    }
}
\caption{Frontier-based exploration}
\label{algo:frontier_policy}
\end{algorithm}

\section{Generating difficult testing episodes for PointNav}
\label{appsec:difficult_pointnav_episodes}

\begin{figure}[h!]
    \centering
    \includegraphics[width=\textwidth]{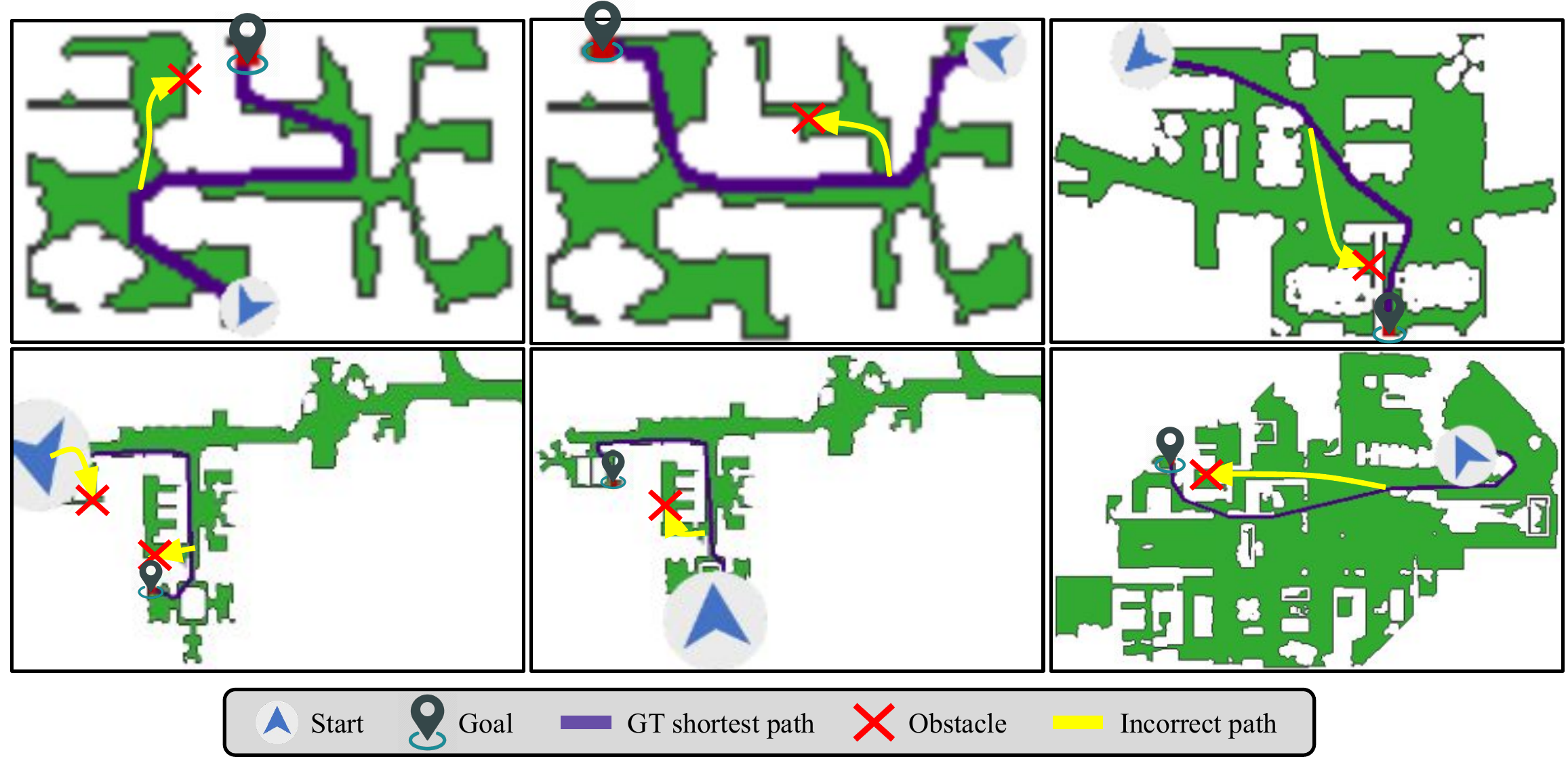}
    \caption{\textbf{Difficult navigation episodes:} Six examples of difficult navigation episodes in MP3D that would benefit from exploration are shown. The ground truth shortest path is shown in purple. The navigation that starts to navigate from ``Start" to ``Goal". If it is not aware of the presence of the obstacle (the red X indicator), it is likely to follow the incorrect yellow path, discover the obstacle and walk back all the way and re-plan. Larger the deviation from the shortest path in purple, lower the navigation efficiency (SPL). Good exploration agents discover these obstacles during exploration and, therefore, have better navigation efficiency.}
    \label{fig:hard_pointnav_episodes}
\end{figure}

In the implementation details provided in Sec.~\ref{sec:experiments} of the main paper, we mentioned that we generate difficult test episodes for navigation.  Here, we describe the rationale behind selecting difficult episodes and show some examples. In order to generate difficult navigation episodes, we ask the following question: 
\begin{equation}
    \textrm{\textit{How difficult would it be for an agent that has not explored the environment to reach the target?}}\nonumber
\end{equation}
An agent that has not explored the environment would assume the entire environment is free and plan accordingly. 
When this assumption fails, i.e., a region that was expected to be free space is blocked, the navigation agent has to most likely reverse course and find a different path (re-plan).
As newer obstacles are discovered along the planned shortest paths, navigation efficiency reduces as more re-planning is required. 
Therefore, an exploration agent that uses the exploration time budget to discover these obstacles a priori is expected to have higher navigation efficiency.
We select start and goal points for navigation by manually inspecting floor-plans and identifying candidate (start, goal) locations that will likely require good exploration for efficient navigation. See Fig.~\ref{fig:hard_pointnav_episodes} for some examples on MP3D. We use the same idea to generate episodes for AVD.

\section{Automatically mining landmarks}
\label{appsec:mining_landmarks}
In Sec.~\ref{sec:evaluation_metrics} from the main paper, we briefly motivated what landmarks are and how they are used for learning an exploration policy. Here, we explain how these landmarks are mined automatically from training data. \\

\begin{figure}[t]
    \centering
    \includegraphics[width=1.0\textwidth]{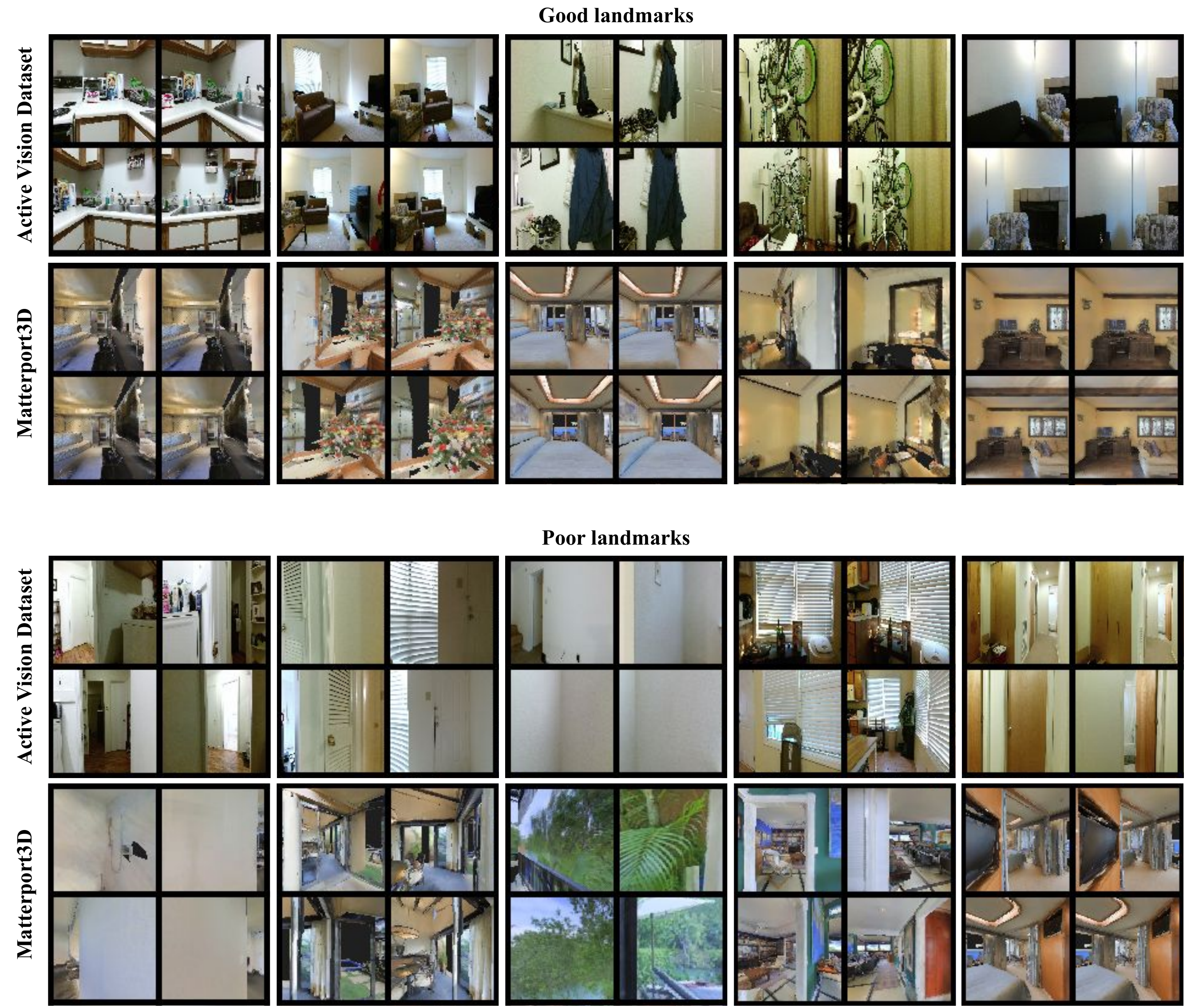}
    \caption{Examples of good and poor landmarks on AVD~\cite{ammirato2016avd}, MP3D~\cite{chang2017matterport} are shown. Good landmarks typically include things that occur uniquely within one environment such as kitchens, living rooms, bedrooms, study tables, etc. Poor landmarks typically include repetitive things in the environment such as doorways, doors, plain walls, and plants. Note that one concept could be a good landmark in one environment, but poor in another. For example, if there is just one television in a house, it is a good landmark. However, as we can see in the last column, last row, televisions that occur more than once in an environment are poor landmarks. } 
    \label{fig:mining_landmarks}
\end{figure}

We define landmarks to be visually distinct parts of the environment, i.e., similar looking viewpoints do not occur in any other spatially distinct part of the environment. To extract such viewpoints from the environment, we sample large number of randomly selected viewpoints from each environment. For each viewpoint, we extract features from a visual similarity prediction network (see Sec.~A.1 in the main paper) and cluster the features using K-Means. Visually similar view-points are clustered together due to the embedding learned by the similarity network. We then sort the clusters based on the intra-cluster variance in the (x, y, z) positions and select clusters with low variance. These clusters include viewpoints which do not have similar views in any other part of the environment, i.e., they are \textit{visually and spatially} distinct. In indoor environments, these typically include distinct objects such as bicycles, mirrors and jackets, and also more abstract concepts such as kitchens, bedrooms and study areas (see Fig.~\ref{fig:mining_landmarks} for examples). 

\begin{table}[ht]
    \begin{minipage}{\linewidth}
        \centering
        \includegraphics[width=\textwidth]{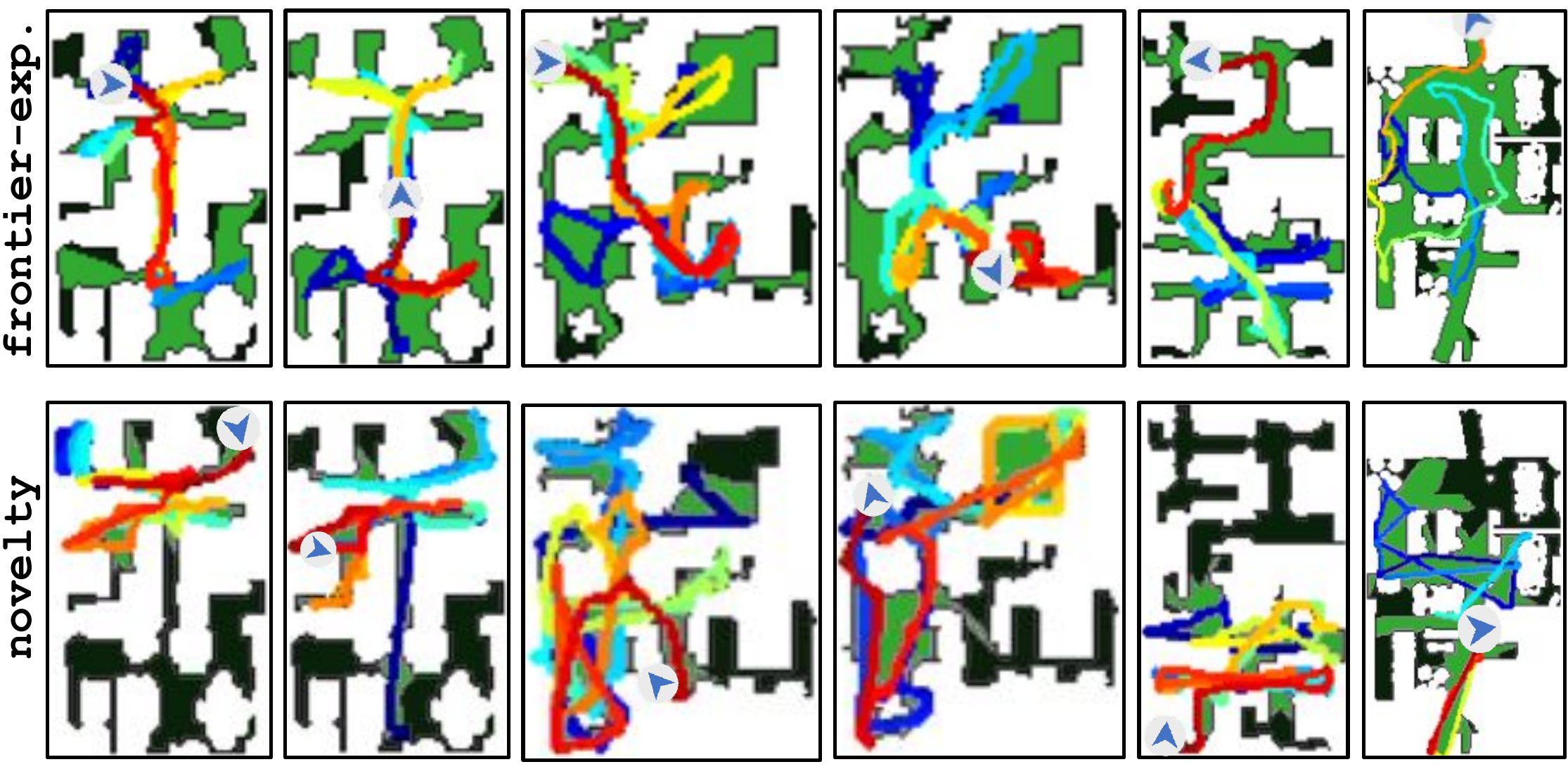}
        \captionof{figure}{\textbf{Success cases of \texttt{frontier-exploration}:} for smaller environments that typically do not have mesh defects, \texttt{\small frontier-exploration} is successful at systematically identifying regions that were not explored and covering them. While \texttt{\small novelty} does fairly well, it generally does worse than \texttt{\small frontier-exploration} on these cases.}
        \label{fig:frontier_succeeds}
    \vspace{0.5cm}
    \end{minipage}
    \begin{minipage}{\linewidth}
        \centering
        \includegraphics[width=\textwidth]{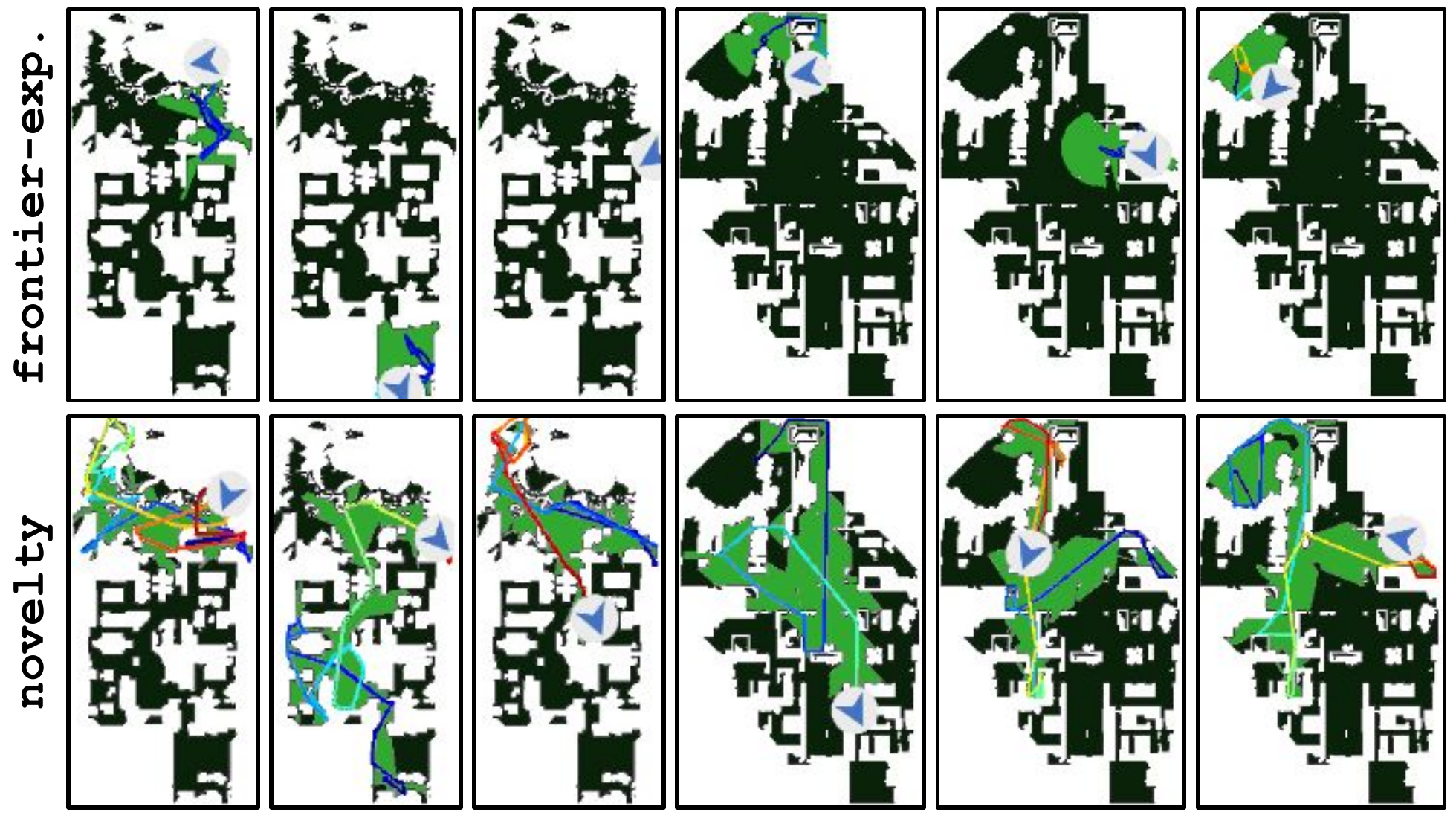}
        \captionof{figure}{\textbf{Failure cases of \texttt{frontier-exploration}:} for larger environments that tend to have either outdoor regions or mesh defects, the occupancy estimation often tends to be incorrect. Since \texttt{frontier-exploration} relies on heuristics for exploration, it is less robust to these noisy cases and gets stuck in regions where noise is high. A learned approach like \texttt{\small novelty} is more robust to these cases.} 
        \label{fig:frontier_fails}
    \end{minipage}
\end{table}

\section{Factors influencing performance}
\label{appsec:factors_influencing}

In Sec. 5.3 from the main paper, we briefly discussed two different factors that affect quality of exploration, specifically, the number of training environments and the size of testing environments. Here, we provide qualitative examples of success and failure cases of \texttt{\small frontier-exploration} in the noise-free case on Matterport3D. We additionally analyze the impact of using imitation learning as a pre-training strategy for learning exploration staretgy.

\subsection{Influence of testing environment size}

As discussed in Sec.~\ref{sec:analysis} in the main paper, \texttt{\small novelty} perform well on most environments. While \texttt{\small frontier-exploration} performs very well in small environments, it struggles in large MP3D environments. This is due to mesh defects present in the scans of large environments where the frontier agent gets stuck. Here, we show qualitative examples where \texttt{\small frontier-exploration} succeeds in small environments (see Fig.~\ref{fig:frontier_succeeds}), and fails in large environments (see Fig.~\ref{fig:frontier_fails}). For each example, we also show the exploration trajectories of \texttt{\small novelty} to serve as a reference because it succeeds on a wide variety of cases (see Fig.~\ref{fig:varying_testing_envs} from the main paper).

\subsection{Influence of imitation-based pre-training}
\begin{figure}[t]
    \centering
    \includegraphics[width=1.0\textwidth]{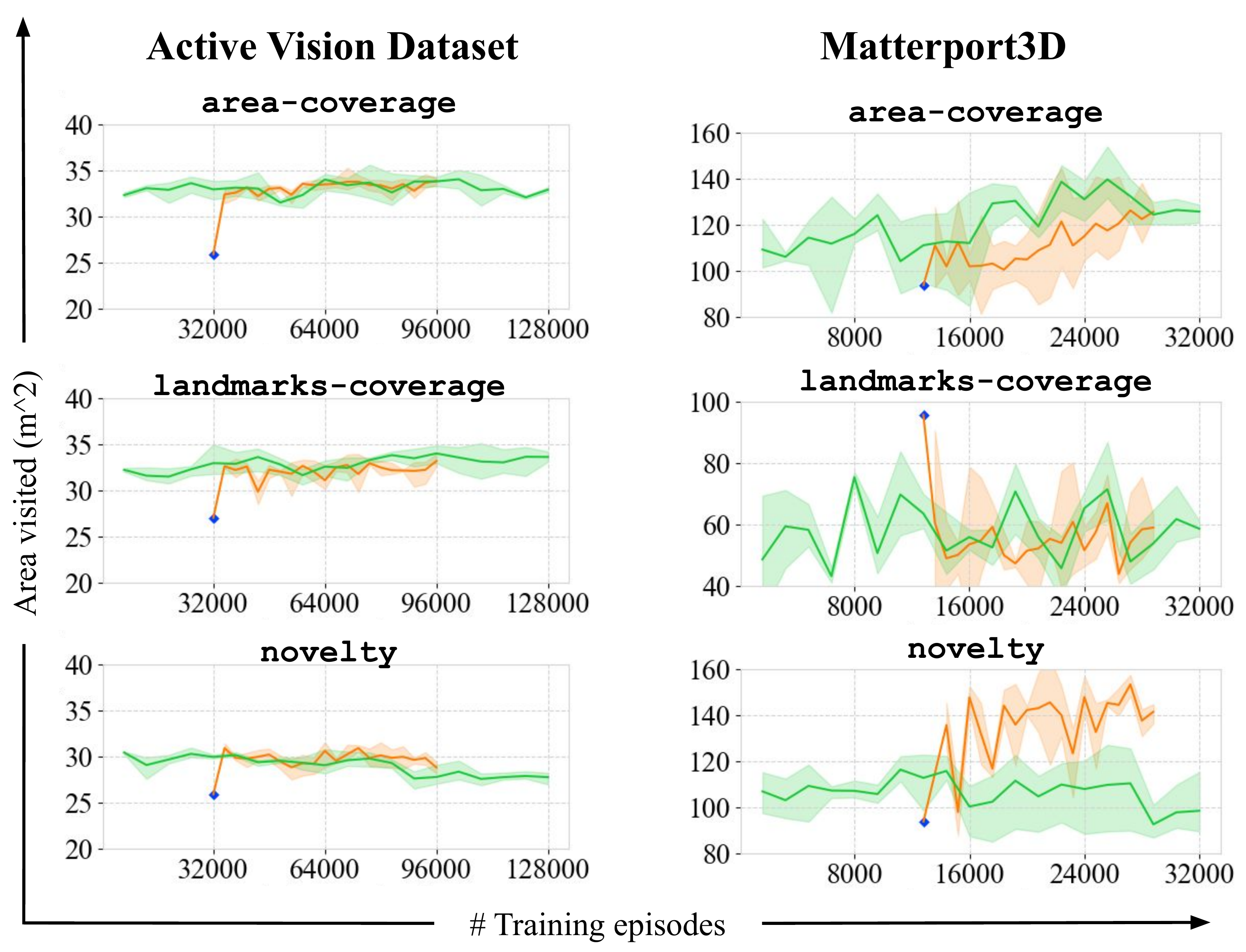}
    \caption{\textbf{Impact of imitation pre-training:} The yellow curves show the results with imitation pre-training, the green curves show results when the policy is trained from random initialization.  The curves represent the area covered by the agent on validation episodes over the period of training. The blue dot indicates the performance of the pure imitation policy. The yellow curves are shifted to account for number of training episodes used for imitation learning.}
    \label{fig:il_analyses}
\end{figure}

In Sec.~\ref{sec:policy_learning_algo} from the main paper, we mentioned that we pre-train policies with imitation learning before the reinforcement learning stage. Here, we evaluate the impact of doing so. See Fig.~\ref{fig:il_analyses}. \\

We train a few sample methods on both AVD and MP3D with, and without the imitation learning stage on three different random seeds. We then evaluate their exploration performance on 100 AVD episodes and 90 MP3D episodes as a function of the number of training episodes. Except in the case of \texttt{\small novelty} in MP3D, pre-training policies using imitation does not seem to improve performance or speed up convergence. While it is possible that expert trajectories gathered from humans (as opposed to synthetically generated) could lead to better performance, we reserve such analyses for future work.

\end{document}